\pdfoutput=1

\documentclass[11pt]{article}

\usepackage[]{ACL2023}

\usepackage{times}
\usepackage{latexsym}

\usepackage[T1]{fontenc}

\usepackage[utf8]{inputenc}

\usepackage{microtype}

\usepackage{inconsolata}
\usepackage{hyperref}
\usepackage{url}
\usepackage{booktabs}
\usepackage{caption}
\usepackage{graphicx}
\usepackage{subfigure}
\usepackage{wrapfig}
\usepackage{multirow}
\usepackage{chngcntr}
\usepackage{mathrsfs}
\usepackage{bbm}
\usepackage{amsmath}
\usepackage{cases}
\usepackage{colortbl}
\usepackage{tcolorbox}
\tcbuselibrary{breakable}
\usepackage{listings}
\tcbuselibrary{listings}
\usepackage[font=normalsize,skip=5pt]{caption}
\usepackage[ruled,linesnumbered,vlined]{algorithm2e}

\title{Seeker: Towards Exception Safety Code Generation with Intermediate Language Agents Framework}

\author{Xuanming Zhang$^{1,2*}$, Yuxuan Chen$^{1*}$, Yiming Zheng$^{3}$, Zhexin Zhang$^{1}$, Yuan Yuan$^{4}$, Minlie Huang$^{1}$\\
$^{1}$ The CoAI Group, DCST, Tsinghua University\\
$^{2}$ ByteDance \\
$^{3}$ Lingxin AI \\
$^{4}$ Beihang University \\
  \texttt{\{zhangxuanming.1\}@bytedance.com} \\
  \texttt{\{chenyuxu21\}@mails.tsinghua.edu.cn} \\
  \texttt{\{aihuang\}@tsinghua.edu.cn}}

\begin{document}
\maketitle

\begin{abstract}
In real-world software development, improper or missing exception handling can severely impact the robustness and reliability of code. Exception handling mechanisms require developers to detect, capture, and manage exceptions according to high standards, but many developers struggle with these tasks, leading to fragile code. This problem is particularly evident in open-source projects and impacts the overall quality of the software ecosystem. To address this challenge, we explore the use of large language models (LLMs) to improve exception handling in code. Through extensive analysis, we identify three key issues: \textit{Insensitive Detection of Fragile Code}, \textit{Inaccurate Capture of Exception Block}, and \textit{Distorted Handling Solution}. These problems are widespread across real-world repositories, suggesting that robust exception handling practices are often overlooked or mishandled. In response, we propose \textbf{Seeker}, a multi-agent framework inspired by expert developer strategies for exception handling. Seeker uses agents—\textbf{S}canner, D\textbf{e}tector, Pr\textbf{e}dator, Ran\textbf{k}er, and Handl\textbf{er}—to assist LLMs in detecting, capturing, and resolving exceptions more effectively. Our work is the first systematic study on leveraging LLMs to enhance exception handling practices in real development scenarios, providing valuable insights for future improvements in code reliability.  \footnote{Our code is available at \url{https://github.com/XMZhangAI/Seeker}} \footnote{CEE for community-contribution is available at \url{https://common-exception-enumeration.github.io/CEE/}}
\end{abstract}

\renewcommand{\thefootnote}{\fnsymbol{footnote}}
\footnotetext[1]{Equal contribution.}

\section{Introduction}
\label{sec1}

In the era of large language models for code generation (code LLMs) such as DeepSeek-Coder \citep{deepseekcoder}, Code-Llama \citep{codellama}, and StarCoder \citep{starcoder}, ensuring code robustness and security has become paramount alongside functional correctness. Traditional evaluation metrics, like HumanEval \citep{humaneval} which measures the \textit{Pass@k} rate, and repo-level assessments such as CoderEval \citep{codereval} and DevEval \citep{deveval}, primarily focus on the ability of these models to generate correct and functional code based on natural language descriptions and real-world development scenarios. As functional correctness improves, attention has shifted to addressing defects in LLM-generated code \cite{swebench, securityeval, LLM1, gptj}. Notably, KPC \cite{kpc} and Neurex \cite{codebert} explored LLM performance in exception handling, highlighting their potential to predict and mitigate risks before vulnerabilities arise.



\begin{figure}[t]
  \centering
  \includegraphics[width=0.99\linewidth]{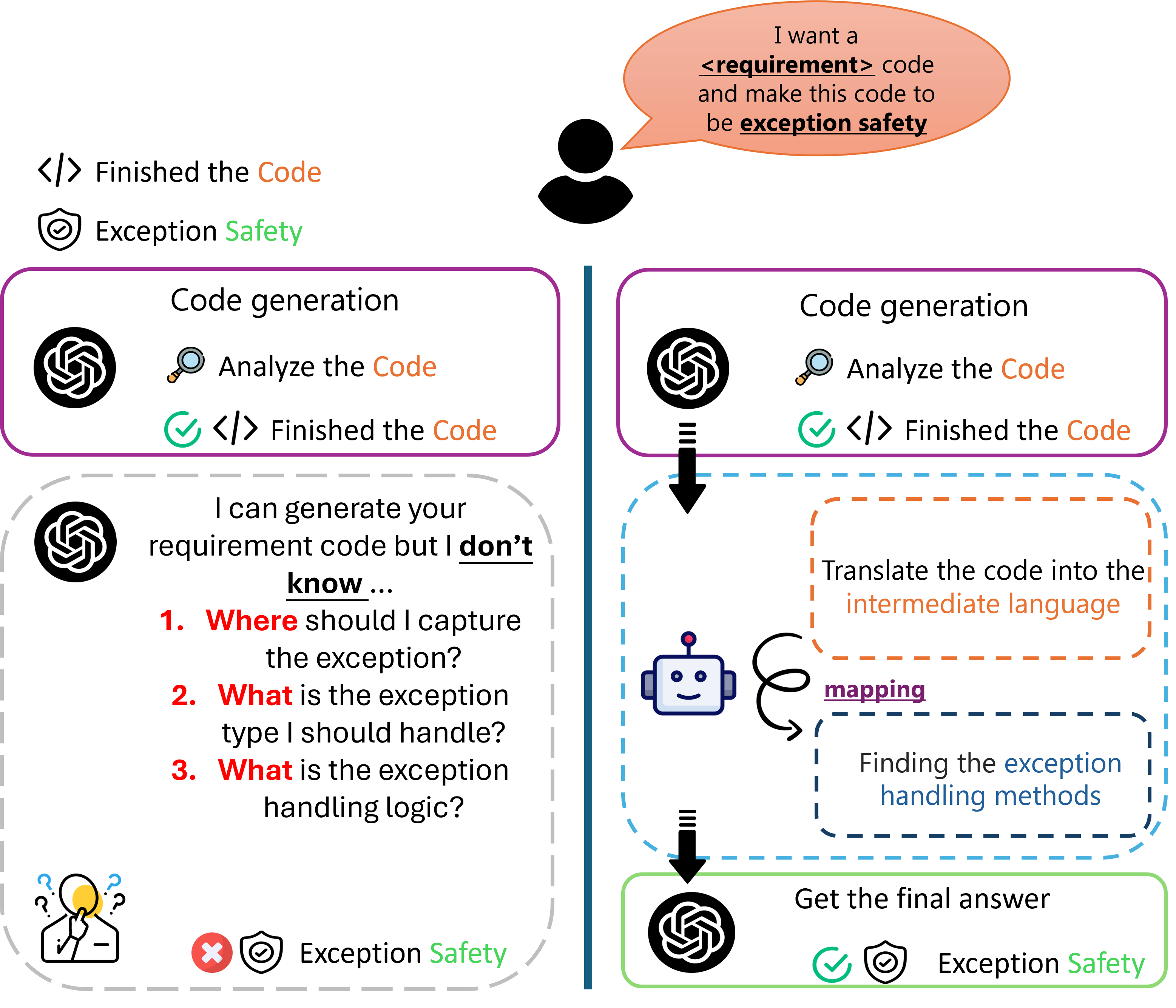}
  \caption{\textbf{ Overview of the Intermediate Language (IL) agents (Right) Compared with Traditional Code Generation Approaches (Left) in Exception-Safe Code Generation Tasks}  The Seeker framework leverages IL agents to perform dynamic analysis, transformation, and optimization of code to ensure robust exception handling. In contrast, traditional approaches often rely on static error-handling routines and lack comprehensive analysis for exception safety.}
  \label{fg1}
  \vskip -.3in
\end{figure}

However, despite advancements in exception detection and handling techniques, there is a significant gap in standardizing exception mechanisms, particularly for custom exceptions and long-tail exception types. Current approaches often define problems narrowly from the perspective of exception handlers, overlooking the potential role of intermediate languages (IL) in managing complex inheritance relationships inherent in exception types. We posit that interpretable and generalizable exception handling strategies are critical yet underestimated factors in real-world code development, profoundly impacting both code robustness and the quality of LLM training data. A schematic diagram of IL Agent can be found at \figurename~\ref{fg1}.

\begin{figure*}
\centering
    \subfigure[Preliminary following the four settings and strategies of KPC. The vertical axis represents the evaluation score of human code review. \label{fig1.1}]{\includegraphics[width=0.45\textwidth]{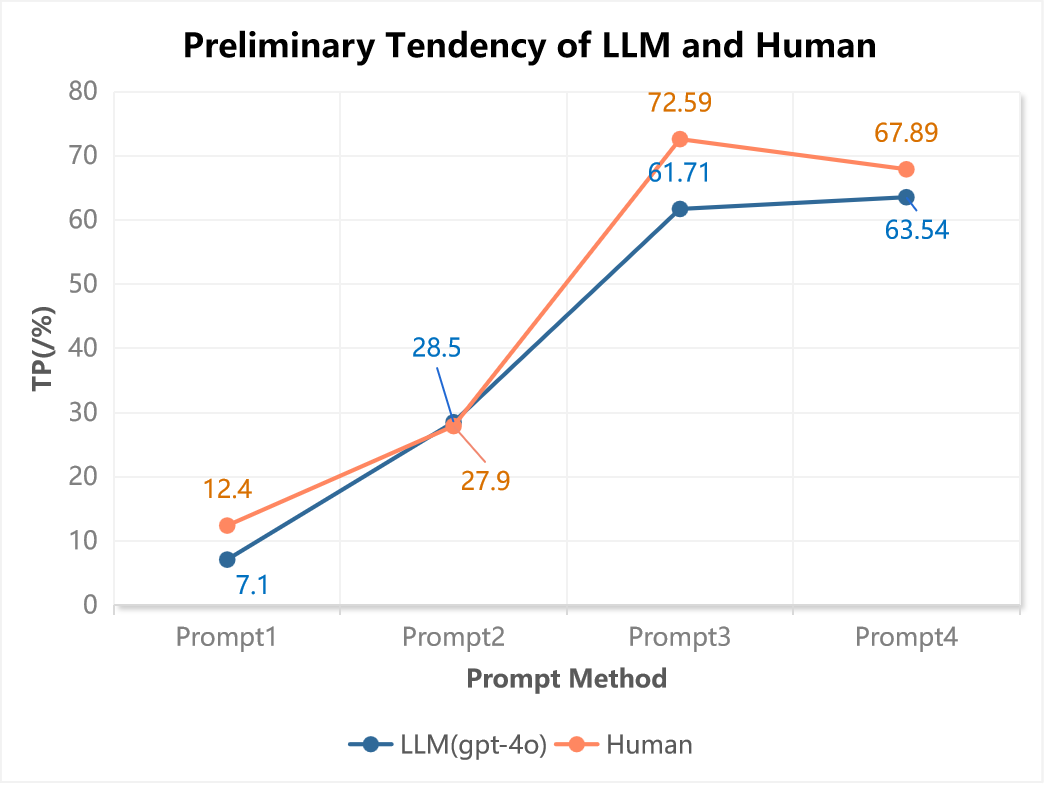}}
    \hfill
    \subfigure[A schematic diagram of human developers who well-performed in exception handling.\label{fig1.2}]{\includegraphics[width=0.45\textwidth]{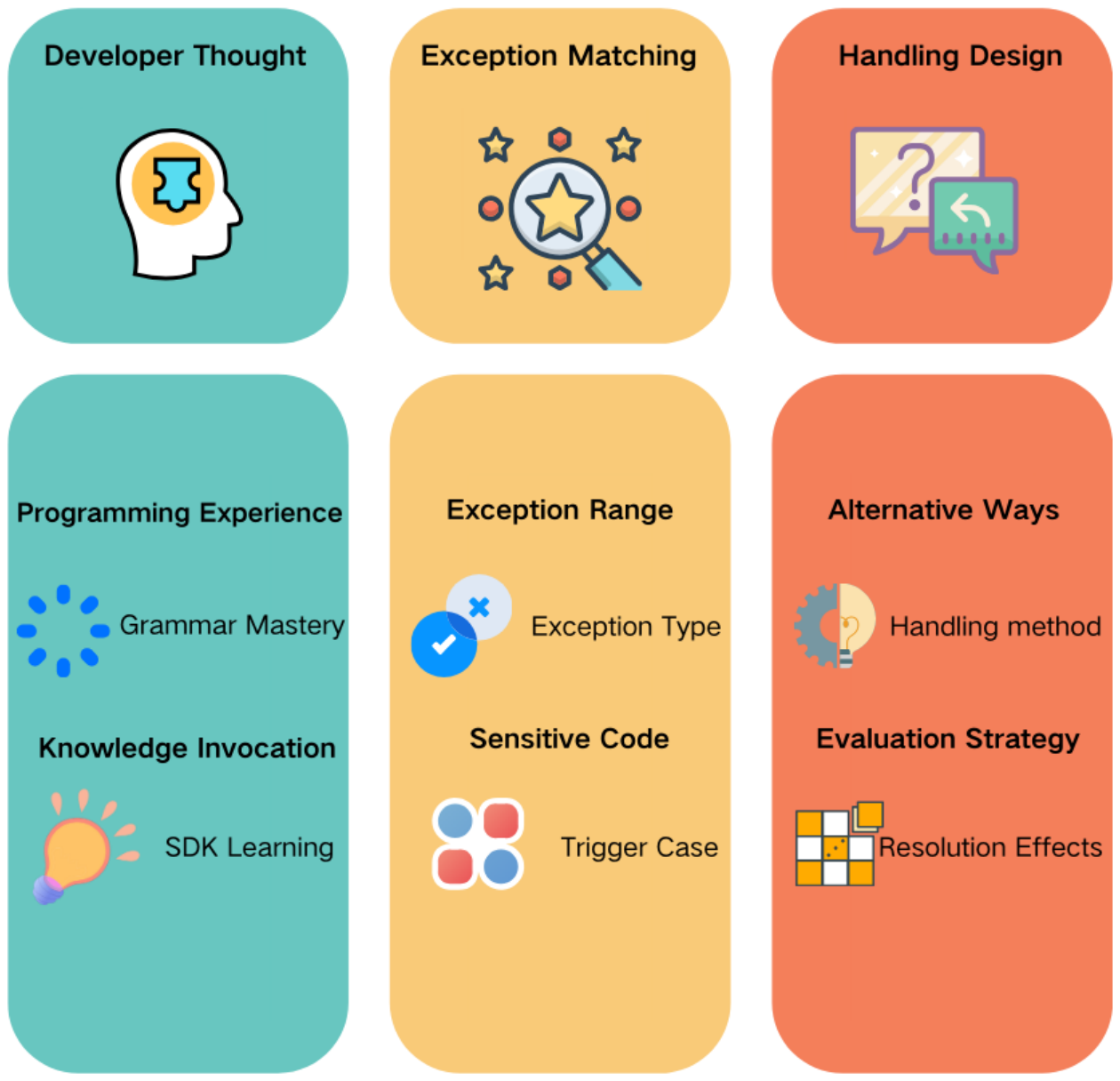}}
\caption{(a) Comparison of LLM and human exception handling performance as prompts evolve from General prompting (Prompt1), to Coarse-grained Knowledge-driven (Prompt2), Fine-grained Knowledge-driven (Prompt3), and Fine-grained Knowledge-driven with explicit handling logic (Prompt4). Results show a clear mitigation effect, where increasingly detailed and context-rich prompts significantly improve handling quality. (b) How expert human developers integrate programming expertise, domain knowledge, fine-grained exception hierarchies, and adaptive strategies to achieve robust exception management.}
\label{fig1}
\vskip -0.1in
\end{figure*}

This paper redefines the research question to address overlooked issues in exception handling by developers, specifically: What method can address developer defects in \textit{Insensitive Detection of Fragile Code}, \textit{Inaccurate Capture of Exception Blocks}, and \textit{Distorted Handling Solutions}? The inquiry highlights the importance of exception safety while exploring how intermediate languages can enhance code analysis beyond traditional human capabilities. To investigate this, we introduce four sets of prompts—Coarse-grained Reminding, Fine-grained Reminding, Fine-grained Inspiring, and Fine-grained Guiding. Experiments validate the effectiveness of fine-grained guiding prompts in improving code exception handling performance. Effective exception handling prioritizes capturing specific exception types, which enables more precise error reporting, such as handling \textit{SQLClientInfoException} instead of its superclass \textit{SQLException} \cite{should-fine}. However, achieving this level of specificity is challenging due to the lack of standardized handling paradigms for long-tail or customized exceptions, and the complexity of intricate inheritance structures.

To enhance code robustness by leveraging best exception handling practices, we propose \textbf{Seeker}, a framework that decomposes exception handling into five specialized tasks managed by distinct agents: \textbf{S}canner: Responsible for scanning the code into manageable unit, D\textbf{e}tector: Responsible for detecting the fragile unit, Pr\textbf{e}dator: Predate the exception block and capture possible exceptions, Ran\textbf{k}er: Sorts exception handling strategies according to certain criteria and selects appropriate exceptions, Handl\textbf{er}: Responsible for performing the final exception handling. We develop Common Exception Enumeration (CEE) from trusted external sources, leveraging explainable IL for exception handling to improve detection, capture, and handling processes where LLMs typically underperform. This approach seamlessly integrates with existing code LLMs to produce highly robust code, while CEE facilitates community contributions by aiding developers in understanding optimal exception handling practices.

Addressing the inefficiency of direct retrieval in complex inheritance trees—exemplified by Java exceptions with 433 nodes, 62 branches, and 5 layers—we introduce a deep retrieval-augmented generation (Deep-RAG) algorithm. This algorithm is tailored to handle intricate inheritance relationships by assigning development scenario labels to branches and employing few-sample verification to fine-tune these labels, thereby enhancing retrieval performance and reducing computational overhead. Experimental results indicate that Seeker significantly improves the robustness and exception handling capabilities of LLM-generated code across various tasks.

In summary, our main contributions are:

\begin{itemize} 
    \setlength{\itemsep}{-5pt}
    \item We highlight the importance of standardization, interpretability, and generalizability in exception handling mechanisms, identifying a critical gap in existing research. 
    \item We propose \textbf{Seeker}, which decomposes exception handling into specialized tasks and incorporates CEE (Contextual Exception Engineering) to enhance performance. 
    \item We conduct extensive experiments demonstrating that \textbf{Seeker} significantly improves code robustness and exception handling performance in LLM-generated code, setting a new state-of-the-art (SOTA) in the field of exception handling. 
\end{itemize}

\section{Preliminary}
\begin{figure*}[t]
  \centering
  \includegraphics[width=0.99\linewidth]{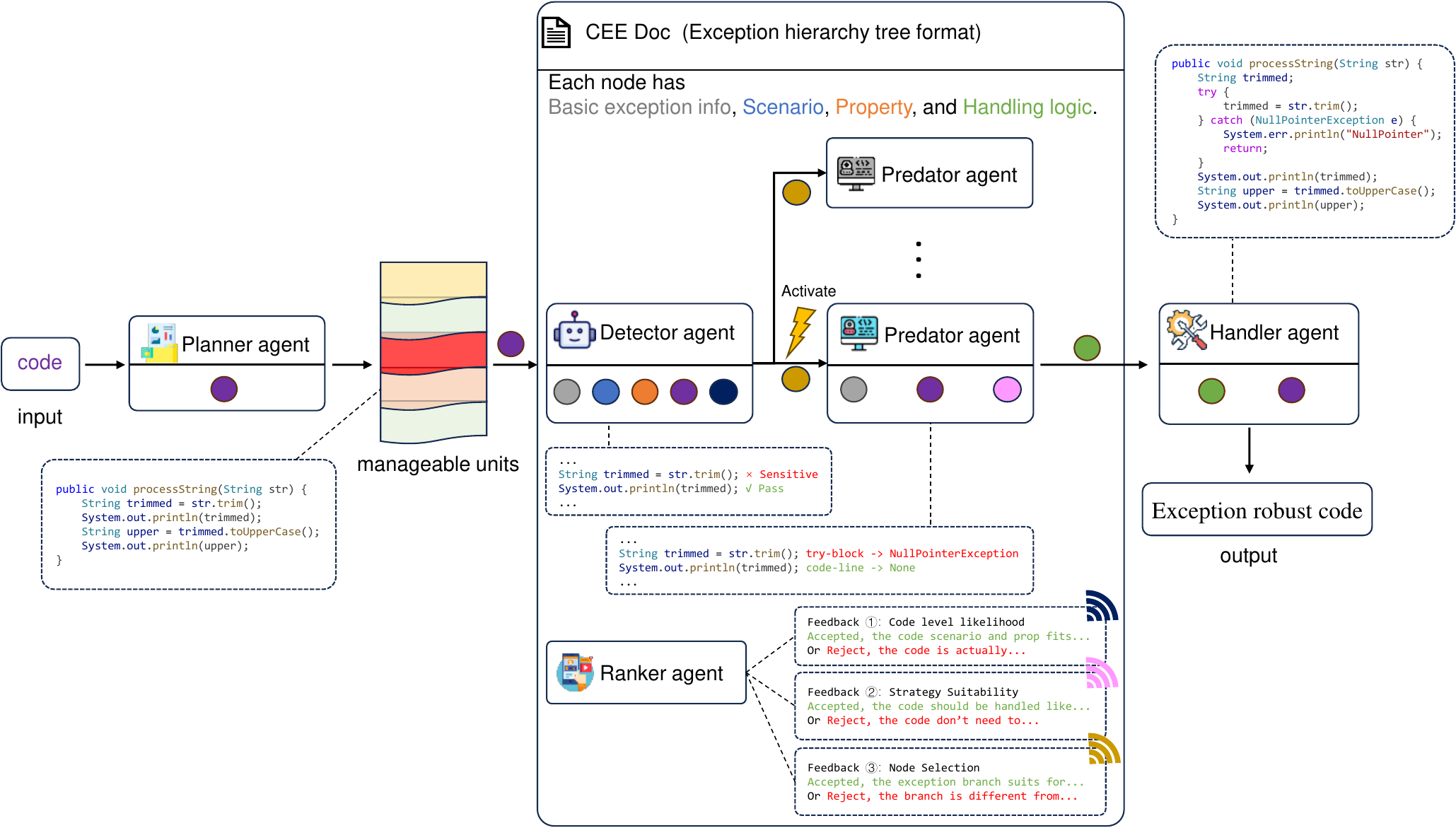}
  \caption{\textbf{Comprehensive Workflow of Seeker.} Seeker orchestrates the automated exception handling process through the seamless collaboration of five specialized agents: Planner, Detector, Predator, Ranker, and Handler. The colored circles within the workflow illustrate the flow of information and interactions among the agents, highlighting how each component activates and contributes to the overall exception handling process. This integrated approach ensures that Seeker delivers highly reliable and maintainable exception handling solutions, significantly improving code robustness and developer productivity.}
  \label{fig4}
  \vskip -.1in
\end{figure*}
\label{sec2.1} This section explores the effectiveness of Intermediate Language (IL)-based Large Language Models in exception handling compared to senior programmers, focusing on standardization, interpretability, and generalizability. Building on prior findings (Appendix \ref{sec3.1.1}), we investigate whether IL can better mitigate exception-handling challenges. Our analysis includes comparative experiments controlling three key factors: standardization of exception types, interpretability of risk scenarios, and generalization of handling strategies. We use four prompt types (Figure \ref{fig2.1} and \ref{fig2.2}): Coarse-grained Reminding, Fine-grained Reminding, Fine-grained Inspiring, and Fine-grained Guiding.

To ensure a practical evaluation, we drew on prior work on KPC\cite{kpc}, selecting well-maintained codebases and using both manual and automated reviews to extract critical exception-handling code. Senior programmers and the IL then analyzed these filtered segments, documenting their processes. To simulate exception-handling thought processes, we established four prompt pathways that progressively offer more detailed guidance, with results illustrated in Figure \ref{fig1.1}.

Our experiments revealed key insights. Prompts lacking clear guidance were ineffective for both IL and programmers. Normative information on exception types helped programmers recognize fragile code but did not significantly improve detection or handling precision. Enhanced scenario interpretability, however, improved understanding and sensitivity to potential fragility, boosting detection accuracy. Generalized handling strategies further improved the analysis of fragility sources, leading to higher-quality exception handling. Together, these enhancements — referred to as the "mitigation effect" — demonstrate that, with proper prompts, IL can match or even surpass senior programmers in exception handling.

These findings inform our proposed $Seeker$ method, which integrates external documentation to generate fine-grained guidance prompts. Appendix \ref{sec3.2.1} offers a deeper analysis of the mitigation effect, supported by data and methodological insights. Figure \ref{fig1.2} illustrates the Chain-of-Thought used by senior developers under Fine-grained Guiding prompts, emphasizing the importance of comprehensive analysis when handling complex exceptions like \textit{BrokenBarrierException} and \textit{AccessControlException}. Our study highlights the potential of IL for reliable code generation and offers a foundation for advancing RAG-based code agents.

\section{Methodology}
\label{headings}
In this section, we introduce the proposed \textbf{Seeker} method, overview in Figure \ref{fig4}. We first review the historical observations of developers on exception handling issues, introducing three exception handling pitfalls, \textit{Insensitive-Detection of Fragile Code}, \textit{Inaccurate-Capture of Exception Block} and \textit{Distorted Handling Solution} in Appendix \ref{sec3.2.1}. Then, we introduce the method's dependency construction and the entire method.

\subsection{Common Exception Enumeration (CEE)} \label{sec3.2.2}

\begin{figure}[t]
  \centering
  \includegraphics[width=0.8\linewidth]{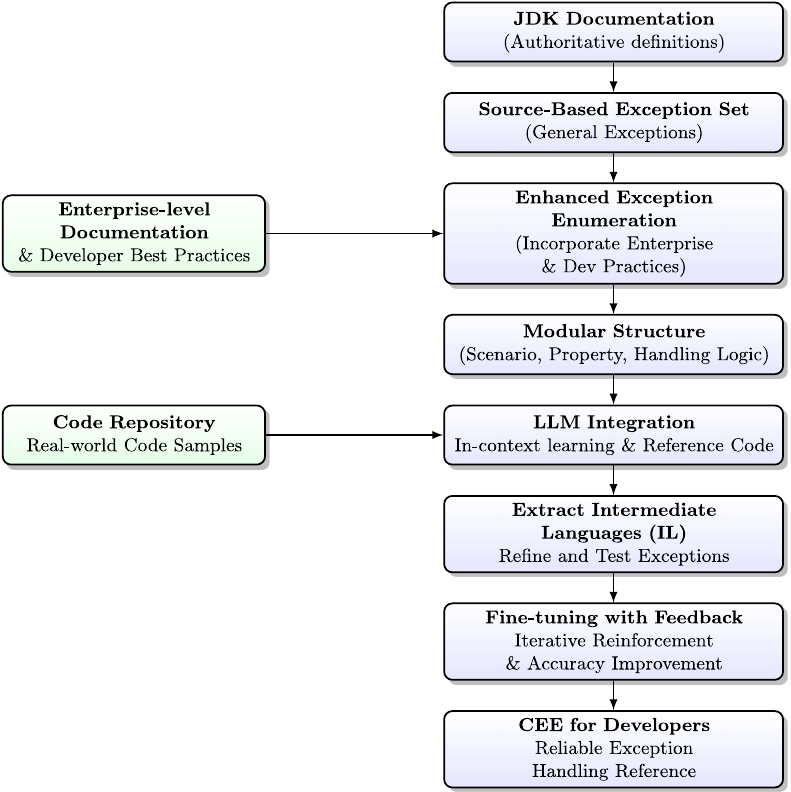}
  \caption{\textbf{An overview of the CEE construction process}. The diagram illustrates how authoritative documentation (JDK), enterprise-level best practices, and real-world code repositories are integrated and refined. Each exception node is enriched with Scenario, Property, and Handling Logic.  This framework is further optimized through LLM-based in-context learning and iterative fine-tuning, ultimately providing a reliable, structured reference (CEE) to enhance exception handling in generated code.}
  \label{fig:cee_schematic}
  \vskip -.3in
\end{figure}

In order to enhance the reliability and robustness of large language model (LLM)-assisted code generation, we introduce the Common Exception Enumeration (CEE). The CEE provides a unified and structured exception documentation base, supporting developers in accurately identifying and handling exceptions. This framework is constructed using authoritative Java Development Kit (JDK) documentation as a foundation and enriched with insights from enterprise practices and open-source code analysis.

The CEE is organized into a hierarchical structure, where each exception node includes three core elements: \textbf{Scenario}, \textbf{Property}, and \textbf{Handling Logic}. By modularizing exception details, developers can more effectively pinpoint the conditions under which exceptions arise, understand their attributes, and apply appropriate handling strategies. This ensures a more proactive and comprehensive approach to exception management than traditional, fragmented methods.

A simplified schematic of the CEE construction process is shown in Figure~\ref{fig:cee_schematic}. This illustration highlights the flow from authoritative documentation (JDK), through curated enterprise insights, to the refined set of exception nodes. Each node is annotated with practical handling logic, enabling more accurate exception responses within generated code.

A more detailed explanation of the construction, iterative refinement processes (including reinforcement learning-based fine-tuning), and community-driven updates is provided in Appendix~\ref{apx:cee_details}.

\subsection{Intermediate Language Agent Framework}
\label{sec:method}

To enhance the standardization, interpretability, and generalizability of exception handling in real-world code development scenarios, we propose a method called \textit{Seeker}. Seeker deconstructs the chain-of-thought processes employed by senior human developers and segments the exception handling mechanism into five specialized tasks, each managed by a dedicated agent: \textit{Planner}, \textit{Detector}, \textit{Predator}, \textit{Ranker}, and \textit{Handler}. By integrating CEE, a substantial repository of trusted external experience documents with IL between exception handling and code, Seeker retrieves and enhances the detection, capture, and handling tasks, addressing areas where the original LLM underperforms. This method can be seamlessly integrated into existing code LLMs to generate highly robust code, with CEE offering valuable community contribution and maintenance benefits, thereby aiding developers in comprehending optimal exception handling practices.

\vspace{-5pt}
\paragraph{Planner}
\textit{Planner} segments the codebase into manageable units such as function blocks, class blocks, and file blocks. This segmentation considers factors like overall code volume, dependency levels, and requirement relationships to mitigate processing pressure, particularly concerning context window limitations and complex dependency chains. By balancing the granularity of segmentation, the Planner ensures that no single unit overwhelms the analysis agents, maintaining clarity and efficiency when handling large and intricate codebases.

\vspace{-5pt}
\paragraph{Detector}
\textit{Detector} concurrently performs scenario and property matching alongside static analysis to identify fragile areas in the code that are susceptible to errors or crashes. Static analysis generates control flow graphs and exception propagation graphs to uncover complex dependencies and deep-level defects, while scenario and property matching captures vulnerabilities based on semantic cues and contextual scenarios that static analysis might overlook. By unifying the results from both methods, the Detector effectively identifies potential exceptions, including long-tail, domain-specific, or customized exception types. However, as discussed in Section \ref{sec1}, detecting exceptions without accounting for complex inheritance relationships may result in inaccurate exception specificity within the hierarchy.

\vspace{-5pt}
\paragraph{Predator}
To address the limitations in exception detection, \textit{Predator} integrates external knowledge from the CEE. Similar to Retrieval-Augmented Generation (RAG) models, Predator summarizes the code at the function level and queries the CEE for relevant exception attributes. It conducts multi-layered deep searches to retrieve applicable information for the detected issues, providing valuable context for exception handling. During few-shot testing phases, feedback on both the accuracy and coverage of the retrieved information is supplied, facilitating the agent’s learning process and enhancing the relevance of future retrievals. We introduce the \textbf{Deep Retrieval-Augmented Generation (Deep-RAG)} algorithm to manage complex inheritance relationships in exception types, as further detailed in Appendix \ref{apx1.1}.

\vspace{-5pt}
\paragraph{Ranker}
\textit{Ranker} assigns grades to detected exceptions based on their likelihood and the suitability of the handling strategies retrieved from the CEE. This grading system prioritizes the most critical exceptions for immediate handling by considering factors such as the probability of occurrence, potential program impact, and the specificity of the exception type within the inheritance hierarchy. The Ranker provides feedback to both the Detector and Predator agents through score ranking and judgment steps, enabling continuous learning from the actual code environment.

\vspace{-5pt}
\paragraph{Handler}
Analyzing the ranked exceptions, \textit{Handler} generates optimized code incorporating robust handling strategies. Utilizing templates and logic patterns derived from the CEE, the Handler ensures that the generated code is functionally correct. It focuses on capturing precise, fine-grained exceptions by navigating down the class hierarchy to provide additional error information beyond superclass exceptions. This approach enhances code readability and maintainability, allowing developers to swiftly identify problem sources and avoid mishandling diverse error types. 

\vspace{-5pt}
\paragraph{Framework}
The general framework for multiple agents is given in Appendix \ref{ag1}.

\vspace{-5pt}
\paragraph{Scalability Considerations}
To mitigate computational overhead, we designed a high-concurrency interface that maintains constant additional computing time overhead regardless of code volume, provided in Appendix \ref{apx2.4}.

\section{Experiments}
\label{experiments}

In this section, we evaluate the performance of our proposed method, \textbf{Seeker}, on the task of exception handling code generation. 


\subsection{Experiment Setup}

\subsubsection{Datasets}
We conduct experiments on a dataset consisting of 750 fragile Java code snippets extracted from real-world projects. These code snippets are selected based on their potential for exception handling improvements, following the rules outlined in Appendix~\ref{apx2.1}.

\subsubsection{Baselines}
We compare \textbf{Seeker} with the following methods:
\vspace{-5pt}
\begin{itemize}
    \setlength{\itemsep}{-5pt}
    \item \textbf{General Prompting}: A straightforward approach where the LLM is prompted to generate exception handling code without any specialized framework or additional knowledge.
    \item \textbf{Traditional Retrieval-Augmented Generation (RAG)}: A method that retrieves relevant information from external sources to assist in code generation.
    \item \textbf{KPC} \citep{kpc}: The state-of-the-art method for exception handling code generation, which leverages knowledge graphs and pattern mining.
    \item \textbf{FuzzyCatch} \citep{baseline3}: A tool for recommending exception handling code for Android Studio based on fuzzy logic.
    \item \textbf{Nexgen} \citep{baseline1}: A neural network pretraining approach for automated exception handling in Java, which predicts try block locations and generates complete catch blocks in relatively high accuracy.
\end{itemize}

For more analysis of related work, see Appendix \ref{rw}.

\subsubsection{Evaluation Metrics}
To comprehensively assess the effectiveness of our method, we employ six quantitative metrics. Details are given in Appendix~\ref{sec:formulations}:
\vspace{-5pt}
\begin{enumerate}
    \setlength{\itemsep}{-5pt}
    \item \textbf{Automated Code Review Score (ACRS)}: Measures the overall adherence to coding standards based on automated code reviews. It provides a percentage indicating how well the code aligns with best practices.
    \item \textbf{Coverage (COV)}: Assesses the proportion of sensitive code segments successfully detected by the system. It reflects the effectiveness of the method in identifying relevant code segments.
    \item \textbf{Coverage Pass (COV-P)}: Focuses on the accuracy of detecting try-blocks. This metric evaluates whether the detected try-blocks match the actual regions requiring exception handling, with penalties for over-detection.
    \item \textbf{Accuracy (ACC)}: Evaluates the correctness of identified exception types. It compares the detected exception types to the actual types, accounting for subclass relationships.
    \item \textbf{Edit Similarity (ES)}: Measures the text similarity between generated and actual try-catch blocks. A higher similarity indicates better quality in the generated code.
    \item \textbf{Code Review Score (CRS)}: Assesses the quality of generated try-catch blocks through evaluation by a language model. This provides a binary evaluation of whether the generated code meets best practices.
\end{enumerate}

\begin{figure*}[!ht]
  \centering
  \includegraphics[width=0.9\linewidth]{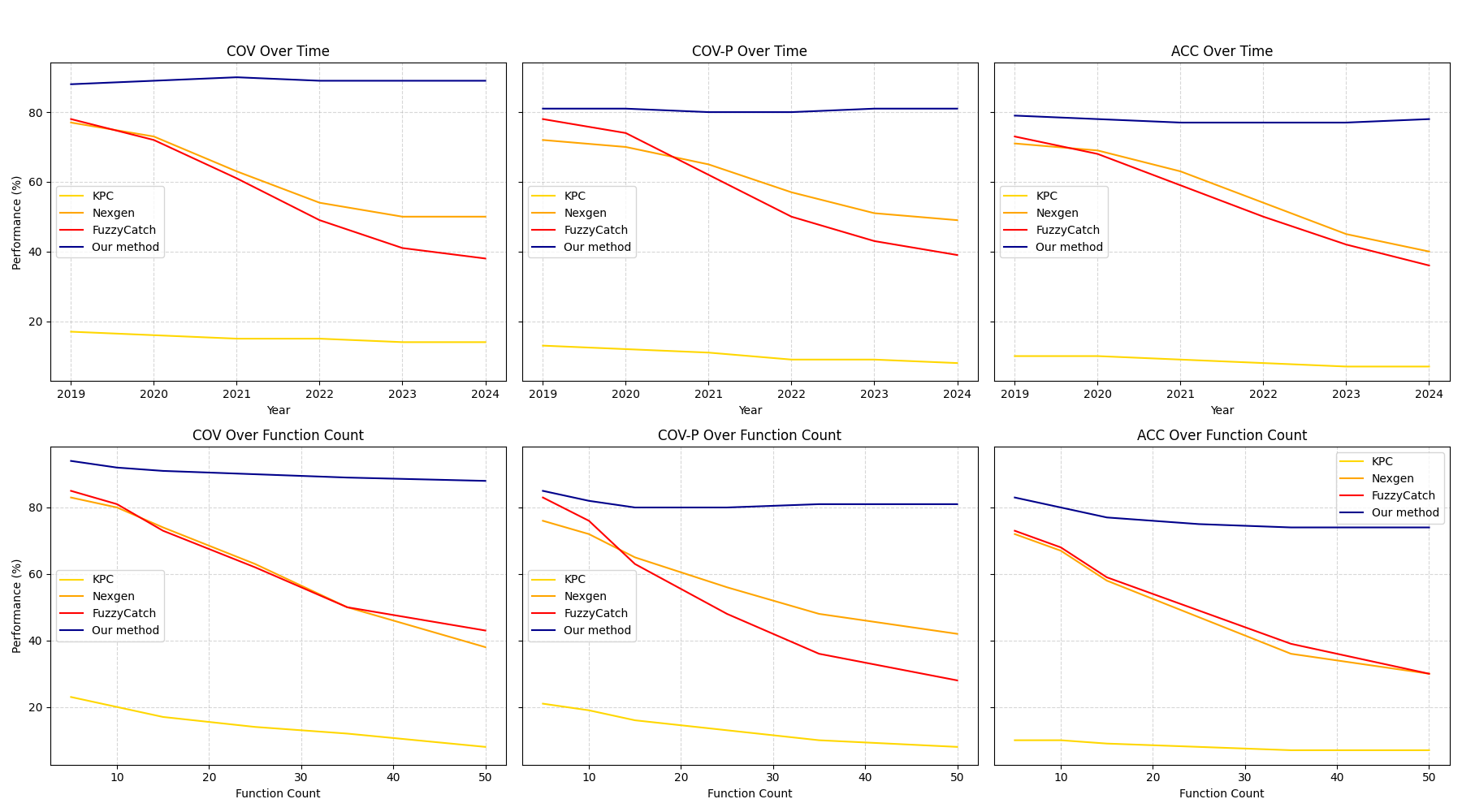}
  \caption{Comparison of Performance Stability Across Baselines and Our Method over Varying Conditions. The top set of curves illustrates the performance metrics over time (2019 to 2024) across different baselines and our method. The bottom set displays performance across increasing function counts.}
  \label{fig8}
  \vskip -.1in
\end{figure*}
\vspace{-5pt}
\subsection{RQ1: Performance Comparison with Baselines}

We compare the performance of \textbf{Seeker} against several baseline methods on the exception handling code generation task. The results are summarized in Table~\ref{table:results}. We conducted our tests on Java code from GitHub spanning a period of five years (2019-2024) and averaged the results across different methods to ensure reliability and account for variability in code quality and project types.

\begin{table}[ht]
\centering
\caption{Comparison of Exception Handling Code Generation Methods}
\label{table:results}
\resizebox{\columnwidth}{!}{  
\begin{tabular}{lcccccc}
\toprule
\textbf{Method} & \textbf{ACRS} & \textbf{COV (\%)} & \textbf{COV-P (\%)} & \textbf{ACC (\%)} & \textbf{ES} & \textbf{CRS (\%)} \\
\midrule
General Prompting       & 0.21 & 13 & 9 & 8 & 0.15 & 24 \\
Traditional RAG          & 0.35 & 35 & 31 & 29 & 0.24 & 31 \\
KPC          & 0.26 & 14 & 11 & 8 & 0.17 & 27 \\
FuzzyCatch & 0.47 & 52 & 50 & 43 & 0.36 & 48 \\
Nexgen & 0.45 & 56 & 49 & 42 & 0.41 & 52 \\
\textbf{Seeker (Ours)}      & \textbf{0.85} & \textbf{91} & \textbf{81} & \textbf{79} & \textbf{0.64} & \textbf{92} \\
\bottomrule
\end{tabular}
}
\vskip -.1in
\end{table}

As depicted in Table~\ref{table:results}, \textbf{Seeker} consistently outperforms all baselines across all evaluation metrics. The key observations include:

\begin{itemize}
    \setlength{\itemsep}{-5pt}
    \item \textbf{ACRS}: \textbf{Seeker} achieves a significantly higher Average Code Review Score (ACRS) of 0.85, indicating superior overall code quality compared to other methods.
    \item \textbf{Coverage (COV) and Coverage Pass (COV-P)}: With COV and COV-P scores of 91\% and 81\%, respectively, \textbf{Seeker} demonstrates exceptional capability in detecting and correctly wrapping sensitive code regions.
    \item \textbf{Accuracy (ACC)}: An ACC of 79\% reflects \textbf{Seeker}'s proficiency in accurately identifying the correct exception types, including recognizing complex subclass relationships.
    \item \textbf{Edit Similarity (ES)}: The ES score of 0.64 indicates that the generated code closely matches the actual exception handling implementations, ensuring minimal discrepancies.
    \item \textbf{Code Review Score (CRS)}: A CRS of 92\% confirms that the code generated by \textbf{Seeker} is highly regarded during automated and LLM-based code reviews, emphasizing its adherence to best practices.
\end{itemize}
\vspace{-5pt}
\paragraph{Depth of Analysis}

Beyond surface-level performance metrics, we explore the relationship between effective exception handling and overall code quality. High ACRS and CRS scores suggest that robust exception handling contributes significantly to maintaining code quality standards. By accurately generating exception handling code, \textbf{Seeker} not only addresses potential runtime issues but also enhances the maintainability and reliability of the software, as reflected in higher code review scores.

\paragraph{Stability and Robustness}

Figure~\ref{fig8} illustrates the stability of \textbf{Seeker} compared to baseline methods over time and varying code complexities.

\textbf{Stability Over Time}: \textbf{Seeker} maintains consistently high performance levels across different time periods, whereas baseline methods exhibit significant variability, particularly in recent years. This stability indicates that \textbf{Seeker} is less sensitive to changes in the development environment and can adapt to evolving software trends and requirements.

\textbf{Performance Across Code Complexity}: As the number of functions within test snippets increases, baseline methods show a decline in performance, struggling with higher code complexity. In contrast, \textbf{Seeker} sustains its performance across all levels of complexity, demonstrating its ability to handle intricate codebases effectively.

\subsection{RQ2: Effect of Different Agents in \textbf{Seeker}}

To understand the contribution of each agent within the \textbf{Seeker} framework, we conducted an ablation study by systematically removing one agent at a time. The results are presented in Table~\ref{table:ablation}.

\begin{table}[ht]
\centering
\caption{Ablation Study on the Effect of Different Agents}
\label{table:ablation}
\resizebox{\columnwidth}{!}{
\begin{tabular}{lcccccc}
\toprule
\textbf{Configuration} & \textbf{ACRS} & \textbf{COV (\%)} & \textbf{COV-P (\%)} & \textbf{ACC (\%)} & \textbf{ES} & \textbf{CRS (\%)} \\
\midrule
\textbf{Seeker (Full)}         & \textbf{0.85} & \textbf{91} & \textbf{81} & \textbf{79} & \textbf{0.64} & \textbf{92} \\
- Scanner       & 0.78 & 85 & 75 & 73 & 0.59 & 86 \\
- Detector      & 0.76 & 63 & 54 & 61 & 0.51 & 84 \\
- Predator      & 0.72 & 61 & 53 & 42 & 0.47 & 81 \\
- Ranker       & 0.63 & 90 & 79 & 75 & 0.49 & 65 \\
- Handler      & 0.50 & 91 & 81 & 79 & 0.34 & 42 \\
\bottomrule
\end{tabular}
}
\vspace{-4ex}
\end{table}

\begin{itemize}
    \setlength{\itemsep}{-5pt}
    \item \textbf{Scanner}: Removing the Scanner results in a notable decrease in ACRS and CRS, underscoring its vital role in initial code analysis and overall quality assessment.
    \item \textbf{Detector}: Its absence significantly reduces coverage metrics, highlighting its importance in identifying sensitive code regions.
    \item \textbf{Predator}: The Predator is essential for accurate exception type detection, as evidenced by the sharp decline in ACC and related metrics when it is removed.
    \item \textbf{Ranker}: Without the Ranker, the selection of handling strategies becomes less effective, impacting the edit similarity and code review scores.
    \item \textbf{Handler}: The most substantial drop in CRS occurs without the Handler, indicating its critical role in implementing exception handling correctly.
\end{itemize}
The ablation study reveals that each agent in the \textbf{Seeker} framework contributes uniquely to its overall performance. The interplay between agents ensures comprehensive exception handling, from initial detection to the implementation of robust handling strategies. This synergy not only enhances exception handling but also positively impacts code quality reviews, as robust exception handling is a key indicator of well-structured and maintainable code.



\subsection{RQ3: Effect of Underlying Language Model}

To evaluate the impact of the underlying LLM on \textbf{Seeker}'s performance, we implemented \textbf{Seeker} using different models, including various open-source and close-source models. The detailed results are provided in Appendix \ref{apx2.6}.

Advanced language models like GPT-4o significantly enhance \textbf{Seeker}'s performance across all metrics. This improvement suggests that the capabilities of the underlying LLM, such as understanding complex code structures and generating accurate exception handling code, play a crucial role in the overall effectiveness of \textbf{Seeker}.

\subsection{RQ4: Impact of Domain-Specific Knowledge Integration}

To assess the impact of integrating domain-specific knowledge, we compared \textbf{Seeker} with and without the inclusion of the CEE. The results are presented in Table~\ref{table:cee}.

\begin{table}[h]
\centering
\caption{Impact of Integrating Common Exception Enumeration (CEE)}
\label{table:cee}
\resizebox{\columnwidth}{!}{
\begin{tabular}{lcccccc}
\toprule
\textbf{Configuration} & \textbf{ACRS} & \textbf{COV (\%)} & \textbf{COV-P (\%)} & \textbf{ACC (\%)} & \textbf{ES} & \textbf{CRS (\%)} \\
\midrule
\textbf{Seeker}     & \textbf{0.85} & \textbf{91} & \textbf{81} & \textbf{79} & \textbf{0.64} & \textbf{92} \\
- CEE       & 0.38 & 48 & 41 & 32 & 0.29 & 46 \\
\bottomrule
\end{tabular}
}
\vskip -.2in
\end{table}

The inclusion of CEE results in substantial improvements across all metrics. Specifically, ACRS increases from 0.38 to 0.85, and CRS jumps from 46\% to 92\%. This significant enhancement highlights the importance of domain-specific knowledge in accurately detecting and handling exceptions, thereby improving overall code quality and reliability.

It is worth mentioning that even without the CEE document, the CoT process for the model under the Seeker framework has brought effective improvement compared to general prompting, which illustrates the effectiveness of the Seeker program analysis framework.

\subsection{Additional Analysis}

Beyond the primary research questions, we conducted additional experiments to evaluate \textbf{Seeker}'s performance in generating repository-level code and optimizing code patches for GitHub issues. Detailed results are available in Appendix~\ref{apx2.5}.

\textbf{Seeker} maintains competitive performance in these real-world scenarios, demonstrating its robustness and applicability. The ability to generate repository-level code and effectively optimize code patches underscores \textbf{Seeker}'s versatility and its potential to assist developers in maintaining high-quality codebases.

\section{Conclusion}
\label{sec:bibtex}

In this paper, we extend the study of the impact of prompt specifications on the robustness of LLM generated code. We conduct extensive comparative experiments using four sets of prompt settings and further confirm the mitigating effect of developers' poor exception handling practices. To exploit this phenomenon, we introduce the Seeker method, a multi-agent collaboration framework that provides LLM with the prompt information required for mitigation effects with the support of CEE documents and Deep-RAG algorithms. The upper bound model achieves SOTA performance on exception handling tasks. In general, Seeker can be integrated into any base model, extended to multiple programming languages, and even generalized to knowledge analysis and reasoning of general inheritance relations, such as requirements engineering in Appendix \ref{apx2.5}. We hope that our findings and proposed methods can provide new insights and promote future research in these areas.

\section{Limitations}
\paragraph{Base Model} The actual performance of Seeker framework depends on the base model's code understanding and information analysis capabilities. Therefore, in order to achieve good experimental performance, it is necessary to introduce base model with strong general capabilities.

\paragraph{Closed-source Model} The good performance of closed-source models may lead to more overhead and privacy leakage issues in enterprise application-level development.

\bibliography{anthology,custom}

\begin{thebibliography}{50}
\expandafter\ifx\csname natexlab\endcsname\relax\def\natexlab#1{#1}\fi

\bibitem[{Cai et~al.(2024)Cai, Yadavally, Mishra, Montejo, and Nguyen}]{codebert}
Yuchen Cai, Aashish Yadavally, Abhishek Mishra, Genesis Montejo, and Tien~N. Nguyen. 2024.
\newblock Programming assistant for exception handling with codebert.
\newblock In \emph{ICSE}.

\bibitem[{Chen et~al.(2021)Chen, Tworek, Jun, Yuan, de~Oliveira~Pinto, Kaplan, Edwards, Burda, Joseph, Brockman et~al.}]{humaneval}
Mark Chen, Jerry Tworek, Heewoo Jun, Qiming Yuan, Henrique~Ponde de~Oliveira~Pinto, Jared Kaplan, Harri Edwards, Yuri Burda, Nicholas Joseph, Greg Brockman, et~al. 2021.
\newblock Evaluating large language models trained on code.
\newblock \emph{arXiv preprint arXiv:2107.03374}.

\bibitem[{Clade(2023)}]{claude}
Clade. 2023.
\newblock https://www.anthropic.com/index/claude-2.

\bibitem[{Claus and Boutilier(1998)}]{dy}
Caroline Claus and Craig Boutilier. 1998.
\newblock The dynamics of reinforcement learning in cooperative multiagent systems.
\newblock In \emph{AAAI}.

\bibitem[{Codex(2021)}]{codex}
Codex. 2021.
\newblock https://openai.com/index/openai-codex/.

\bibitem[{de~P{\'{a}}dua and Shang(2017)}]{difficulty1}
Guilherme~B. de~P{\'{a}}dua and Weiyi Shang. 2017.
\newblock Revisiting exception handling practices with exception flow analysis.
\newblock In \emph{SCAM}.

\bibitem[{de~Sousa et~al.(2020)de~Sousa, Maia, Rocha, and Viana}]{difficulty3}
D{\^{e}}mora Bruna~Cunha de~Sousa, Paulo Henrique~M. Maia, Lincoln~S. Rocha, and Windson Viana. 2020.
\newblock Studying the evolution of exception handling anti-patterns in a long-lived large-scale project.
\newblock \emph{J. Braz. Comput. Soc.}

\bibitem[{Dong et~al.(2023)Dong, Jiang, Jin, and Li}]{dong}
Yihong Dong, Xue Jiang, Zhi Jin, and Ge~Li. 2023.
\newblock Self-collaboration code generation via chatgpt.
\newblock \emph{ACM Trans. Softw. Eng. Methodol.}

\bibitem[{Ebert et~al.(2020)Ebert, Castor, and Serebrenik}]{Java}
Felipe Ebert, Fernando Castor, and Alexander Serebrenik. 2020.
\newblock A reflection on "an exploratory study on exception handling bugs in java programs".
\newblock In \emph{SANER}.

\bibitem[{GPT-3(2022)}]{gpt3}
GPT-3. 2022.
\newblock https://platform.openai.com/docs/models/gpt-base.

\bibitem[{GPT-3.5(2023)}]{gpt3.5}
GPT-3.5. 2023.
\newblock https://platform.openai.com/docs/models/\#gpt-3-5-turbo.

\bibitem[{GPT-4(2023)}]{gpt4}
GPT-4. 2023.
\newblock https://platform.openai.com/docs/models/gpt-4.

\bibitem[{GPT-4o(2024)}]{gpt4o}
GPT-4o. 2024.
\newblock https://platform.openai.com/docs/models/gpt-4o.

\bibitem[{Guo et~al.(2024)Guo, Zhu, Yang, Xie, Dong, Zhang, Chen, Bi, Wu, Li et~al.}]{deepseekcoder}
Daya Guo, Qihao Zhu, Dejian Yang, Zhenda Xie, Kai Dong, Wentao Zhang, Guanting Chen, Xiao Bi, Y.~Wu, Y.~K. Li, et~al. 2024.
\newblock Deepseek-coder: When the large language model meets programming - the rise of code intelligence.
\newblock \emph{arXiv preprint arXiv:2401.14196}.

\bibitem[{He and Vechev(2023)}]{LLM1}
Jingxuan He and Martin~T. Vechev. 2023.
\newblock Large language models for code: Security hardening and adversarial testing.
\newblock In \emph{CCS}.

\bibitem[{Huang et~al.(2025)Huang, Zhang, Meng, and Liu}]{a3}
Kai Huang, Jian Zhang, Xiangxin Meng, and Yang Liu. 2025.
\newblock { Template-Guided Program Repair in the Era of Large Language Models }.
\newblock In \emph{ICSE}.

\bibitem[{Jacobs and Piessens(2009)}]{exception3}
Bart Jacobs and Frank Piessens. 2009.
\newblock Failboxes: Provably safe exception handling.
\newblock In \emph{ECOOP}.

\bibitem[{Jimenez et~al.(2024)Jimenez, Yang, Wettig, Yao, Pei, Press, and Narasimhan}]{swebench}
Carlos~E. Jimenez, John Yang, Alexander Wettig, Shunyu Yao, Kexin Pei, Ofir Press, and Karthik~R. Narasimhan. 2024.
\newblock Swe-bench: Can language models resolve real-world github issues?
\newblock In \emph{ICLR}.

\bibitem[{Li et~al.(2023{\natexlab{a}})Li, Hammoud, Itani, Khizbullin, and Ghanem}]{camel}
Guohao Li, Hasan Abed Al~Kader Hammoud, Hani Itani, Dmitrii Khizbullin, and Bernard Ghanem. 2023{\natexlab{a}}.
\newblock Camel: Communicative agents for "mind" exploration of large language model society.
\newblock In \emph{NeurIPS}.

\bibitem[{Li et~al.(2024{\natexlab{a}})Li, Li, Zhao, Li, Liu, Zhu, Wang, Liu, Fang, Wang et~al.}]{deveval}
Jia Li, Ge~Li, Yunfei Zhao, Yongmin Li, Huanyu Liu, Hao Zhu, Lecheng Wang, Kaibo Liu, Zheng Fang, Lanshen Wang, et~al. 2024{\natexlab{a}}.
\newblock Deveval: {A} manually-annotated code generation benchmark aligned with real-world code repositories.
\newblock In \emph{ACL(Findings)}.

\bibitem[{Li et~al.(2024{\natexlab{b}})Li, Rabbi, Cheng, Sangalay, Tian, and Yang}]{baseline2}
Junjie Li, Fazle Rabbi, Cheng Cheng, Aseem Sangalay, Yuan Tian, and Jinqiu Yang. 2024{\natexlab{b}}.
\newblock An exploratory study on fine-tuning large language models for secure code generation.
\newblock \emph{arXiv preprint 2408.09078}.

\bibitem[{Li et~al.(2024{\natexlab{c}})Li, Sangalay, Cheng, Tian, and Yang}]{gptj}
Junjie Li, Aseem Sangalay, Cheng Cheng, Yuan Tian, and Jinqiu Yang. 2024{\natexlab{c}}.
\newblock Fine tuning large language model for secure code generation.
\newblock In \emph{FORGE}.

\bibitem[{Li et~al.(2024{\natexlab{d}})Li, Zhang, Chen, Liu, Liu, and Chen}]{a4}
Kaixuan Li, Jian Zhang, Sen Chen, Han Liu, Yang Liu, and Yixiang Chen. 2024{\natexlab{d}}.
\newblock Patchfinder: {A} two-phase approach to security patch tracing for disclosed vulnerabilities in open-source software.
\newblock In \emph{ISSTA}.

\bibitem[{Li et~al.(2023{\natexlab{b}})Li, Allal, Zi, Muennighoff, Kocetkov, Mou, Marone, Akiki, Li, Chim et~al.}]{starcoder}
Raymond Li, Loubna~Ben Allal, Yangtian Zi, Niklas Muennighoff, Denis Kocetkov, Chenghao Mou, Marc Marone, Christopher Akiki, Jia Li, Jenny Chim, et~al. 2023{\natexlab{b}}.
\newblock Starcoder: may the source be with you!
\newblock \emph{TMLR}.

\bibitem[{Li et~al.(2023{\natexlab{c}})Li, Ren, Xue, Xing, and Sun}]{final}
Xiangwei Li, Xiaoning Ren, Yinxing Xue, Zhenchang Xing, and Jiamou Sun. 2023{\natexlab{c}}.
\newblock Prediction of vulnerability characteristics based on vulnerability description and prompt learning.
\newblock In \emph{SANER}.

\bibitem[{Luo et~al.(2024)Luo, Xu, Zhao, Sun, Geng, Hu, Tao, Ma, Lin, and Jiang}]{wizardcoder}
Ziyang Luo, Can Xu, Pu~Zhao, Qingfeng Sun, Xiubo Geng, Wenxiang Hu, Chongyang Tao, Jing Ma, Qingwei Lin, and Daxin Jiang. 2024.
\newblock Wizardcoder: Empowering code large language models with evol-instruct.
\newblock In \emph{ICLR}.

\bibitem[{Minsky(2007)}]{book}
Marvin Minsky. 2007.
\newblock The emotion machine: Commonsense thinking, artificial intelligence, and the future of the human mind.
\newblock \emph{Simon and Schuster}.

\bibitem[{Nakshatri et~al.(2016)Nakshatri, Hegde, and Thandra}]{exception1}
Suman Nakshatri, Maithri Hegde, and Sahithi Thandra. 2016.
\newblock Analysis of exception handling patterns in java projects: an empirical study.
\newblock In \emph{MSR}.

\bibitem[{Nguyen et~al.(2020{\natexlab{a}})Nguyen, Vu, and Nguyen}]{baseline3}
Tam Nguyen, Phong Vu, and Tung Nguyen. 2020{\natexlab{a}}.
\newblock Code recommendation for exception handling.
\newblock In \emph{ESEC/FSE}.

\bibitem[{Nguyen et~al.(2020{\natexlab{b}})Nguyen, Vu, and Nguyen}]{difficulty2}
Tam Nguyen, Phong Vu, and Tung Nguyen. 2020{\natexlab{b}}.
\newblock Code recommendation for exception handling.
\newblock In \emph{ESEC/FSE}.

\bibitem[{o1(2024)}]{o1}
OpenAI o1. 2024.
\newblock https://platform.openai.com/docs/models/o1.

\bibitem[{Osman et~al.(2017)Osman, Chis, Schaerer, Ghafari, and Nierstrasz}]{should-fine}
Haidar Osman, Andrei Chis, Jakob Schaerer, Mohammad Ghafari, and Oscar Nierstrasz. 2017.
\newblock On the evolution of exception usage in java projects.
\newblock In \emph{SANER}.

\bibitem[{Ren et~al.(2023)Ren, Ye, Zhao, Xing, and Yang}]{kpc}
Xiaoxue Ren, Xinyuan Ye, Dehai Zhao, Zhenchang Xing, and Xiaohu Yang. 2023.
\newblock From misuse to mastery: Enhancing code generation with knowledge-driven {AI} chaining.
\newblock In \emph{ASE}.

\bibitem[{Rozi{\`{e}}re et~al.(2023)Rozi{\`{e}}re, Gehring, Gloeckle, Sootla, Gat, Tan, Adi, Liu, Remez, Rapin et~al.}]{codellama}
Baptiste Rozi{\`{e}}re, Jonas Gehring, Fabian Gloeckle, Sten Sootla, Itai Gat, Xiaoqing~Ellen Tan, Yossi Adi, Jingyu Liu, Tal Remez, J{\'{e}}r{\'{e}}my Rapin, et~al. 2023.
\newblock Code llama: Open foundation models for code.
\newblock \emph{arXiv preprint arXiv:2308.12950}.

\bibitem[{Shen et~al.(2023)Shen, Song, Tan, Li, Lu, and Zhuang}]{hug}
Yongliang Shen, Kaitao Song, Xu~Tan, Dongsheng Li, Weiming Lu, and Yueting Zhuang. 2023.
\newblock Hugginggpt: Solving {AI} tasks with chatgpt and its friends in huggingface.
\newblock \emph{NeurIPS}.

\bibitem[{Siddiq and Santos(2022)}]{securityeval}
Mohammed~Latif Siddiq and Joanna C.~S. Santos. 2022.
\newblock Securityeval dataset: Mining vulnerability examples to evaluate machine learning-based code generation techniques.
\newblock In \emph{MSR4P\&S}.

\bibitem[{Smoliar(1991)}]{mind}
Stephen~W. Smoliar. 1991.
\newblock Marvin minsky, the society of mind.
\newblock \emph{Artif. Intell.}, 48(3):349--370.

\bibitem[{Tao et~al.(2024)Tao, Zhou, Zhang, and Cheng}]{a2}
Wei Tao, Yucheng Zhou, Wenqiang Zhang, and Yu~Cheng. 2024.
\newblock {MAGIS:} llm-based multi-agent framework for github issue resolution.
\newblock \emph{arXiv preprint 2403.17927}.

\bibitem[{Wang et~al.(2024)Wang, Jiang, Liu, Chen, and Zheng}]{LLM2}
Yanlin Wang, Tianyue Jiang, Mingwei Liu, Jiachi Chen, and Zibin Zheng. 2024.
\newblock Beyond functional correctness: Investigating coding style inconsistencies in large language models.
\newblock \emph{arXiv preprint 2407.00456}.

\bibitem[{Weimer and Necula(2004)}]{exception2}
Westley Weimer and George~C. Necula. 2004.
\newblock Finding and preventing run-time error handling mistakes.
\newblock In \emph{OOPSLA}.

\bibitem[{Wen et~al.(2023)Wen, Chen, Gao, Zhang, Zhang, and Liao}]{v1}
Xin{-}Cheng Wen, Yupan Chen, Cuiyun Gao, Hongyu Zhang, Jie~M. Zhang, and Qing Liao. 2023.
\newblock Vulnerability detection with graph simplification and enhanced graph representation learning.
\newblock In \emph{ICSE}.

\bibitem[{Wu et~al.(2023)Wu, Yin, Qi, Wang, Tang, and Duan}]{visual}
Chenfei Wu, Shengming Yin, Weizhen Qi, Xiaodong Wang, Zecheng Tang, and Nan Duan. 2023.
\newblock Visual chatgpt: Talking, drawing and editing with visual foundation models.
\newblock \emph{arXiv preprint 2303.04671}.

\bibitem[{Yang et~al.(2024)Yang, Jimenez, Wettig, Lieret, Yao, Narasimhan, and Press}]{yang2024sweagentagentcomputerinterfacesenable}
John Yang, Carlos~E. Jimenez, Alexander Wettig, Kilian Lieret, Shunyu Yao, Karthik Narasimhan, and Ofir Press. 2024.
\newblock Swe-agent: Agent-computer interfaces enable automated software engineering.

\bibitem[{Yu et~al.(2024)Yu, Shen, Ran, Zhang, Zhang, Ma, Liang, Li, Wang, and Xie}]{codereval}
Hao Yu, Bo~Shen, Dezhi Ran, Jiaxin Zhang, Qi~Zhang, Yuchi Ma, Guangtai Liang, Ying Li, Qianxiang Wang, and Tao Xie. 2024.
\newblock Codereval: {A} benchmark of pragmatic code generation with generative pre-trained models.
\newblock In \emph{ICSE}.

\bibitem[{Zhang et~al.(2023)Zhang, Luo, Hu, Yan, Zhang, and Qiu}]{bugs}
Hao Zhang, Ji~Luo, Mengze Hu, Jun Yan, Jian Zhang, and Zongyan Qiu. 2023.
\newblock Detecting exception handling bugs in {C++} programs.
\newblock In \emph{ICSE}.

\bibitem[{Zhang et~al.(2020)Zhang, Wang, Zhang, Sun, Pu, and Liu}]{baseline1}
Jian Zhang, Xu~Wang, Hongyu Zhang, Hailong Sun, Yanjun Pu, and Xudong Liu. 2020.
\newblock Learning to handle exceptions.
\newblock In \emph{ASE}.

\bibitem[{Zhang et~al.(2024{\natexlab{a}})Zhang, Li, Li, Shi, and Jin}]{zhang}
Kechi Zhang, Jia Li, Ge~Li, Xianjie Shi, and Zhi Jin. 2024{\natexlab{a}}.
\newblock Codeagent: Enhancing code generation with tool-integrated agent systems for real-world repo-level coding challenges.
\newblock In \emph{ACL}.

\bibitem[{Zhang et~al.(2024{\natexlab{b}})Zhang, Yuan, and Yao}]{dot}
Yifan Zhang, Yang Yuan, and Andrew Chi-Chih Yao. 2024{\natexlab{b}}.
\newblock On the diagram of thought.
\newblock \emph{arXiv preprint 2409.10038}.

\bibitem[{Zheng et~al.(2023)Zheng, Chiang, Sheng, Zhuang, Wu, Zhuang, Lin, Li, Li, Xing et~al.}]{vicuna}
Lianmin Zheng, Wei{-}Lin Chiang, Ying Sheng, Siyuan Zhuang, Zhanghao Wu, Yonghao Zhuang, Zi~Lin, Zhuohan Li, Dacheng Li, Eric~P. Xing, et~al. 2023.
\newblock Judging llm-as-a-judge with mt-bench and chatbot arena.
\newblock In \emph{NeurIPS}.

\bibitem[{Zhou et~al.(2012)Zhou, Zhang, and Lo}]{a1}
Jian Zhou, Hongyu Zhang, and David Lo. 2012.
\newblock Where should the bugs be fixed? more accurate information retrieval-based bug localization based on bug reports.
\newblock In \emph{ICSE}.

\end{thebibliography}
\bibliographystyle{acl_natbib}

\clearpage
\appendix

\section{Appendix}
\label{sec:appendix}

\subsection{A Revisit of Human Empiricals} \label{sec3.1.1}
Over the years, there have been numerous empirical studies and practical discussions on exception handling, but what is common is that exception handling has been repeatedly emphasized as an important mechanism directly related to code robustness. Exception handling is a necessary and powerful mechanism to distinguish error handling code from normal code, so that the software can do its best to run in a normal state\cite{exception1}. Exception mechanism ensures that unexpected errors do not damage the stability or security of the system, prevents resource leakage, ensures data integrity, and ensures that the program still runs correctly when unforeseen errors occur\cite{exception2}. In addition, exception handling also involves solving potential errors in the program flow, which can mitigate or eliminate defects that may cause program failure or unpredictable behavior\cite{exception3}.

Although the exception mechanism is an important solution to code robustness, developers have always shown difficulties in dealing with it due to its complex inheritance relationship and processing methods. Various programming language projects show a long-tail distribution of exception types when facing exception handling, which means that developers may only have a simple understanding of the frequently occurring exception types\cite{difficulty1}. However, according to section\ref{sec1}, good exception practices rely on developers to perform fine-grained specific capturing. Multi-pattern effect of exception handling is also considerable\cite{difficulty2}. For example, even for peer code, capturing different exception types will play different maintenance functions, so exception handling is often not generalized or single-mapped. These complex exception mechanism practice skills have high requirements for developers' programming literacy. Previous study manually reviewed and counted the exception handling of a large number of open source projects, and believed that up to 62.91\% of the exception handling blocks have violations such as capturing general exceptions and destructive wrapping\cite{difficulty3}. This seriously violates the starting point of the exception mechanism. We also emphasize the urgent need and importance of automated exception handling suggestion tools\cite{difficulty1}.

The failure of human developers in the exception handling mechanism seriously affects the quality of LLM's code training data \cite{LLM1}, which further leads to LLM's inability to understand the usage skills of maintenance functions \cite{LLM2}. To solve the above problems, we first proposed $Seeker-Java$ for the Java language. This is because the Java language has a more urgent need for exception handling and is completely mapped to the robustness of Java programs. As a fully object-oriented language, Java's exception handling is more complex than other languages, and it has a higher degree of integration into language structures\cite{Java}. Therefore, Java projects are more seriously troubled by exception handling bugs. In addition, Java relies heavily on exceptions as a mechanism for handling exceptional events. In contrast, other languages may use different methods or have less strict exception handling mechanisms. It is worth mentioning that $Seeker$'s collaborative solution based on an inherent multi-agent framework plus an external knowledge base, they can quickly migrate multiple languages by maintaining documents for different languages. We will also maintain $Seeker-Python$ and $Seeker-C\#$ in the future to provide robustness guarantees for the development of more programming languages.

\subsection{Method Details} \label{apx1}

\subsubsection{Rules of Good Practice} \label{sec3.2.1}

\begin{figure*}[ht]
  \centering
  \includegraphics[width=0.69\linewidth]{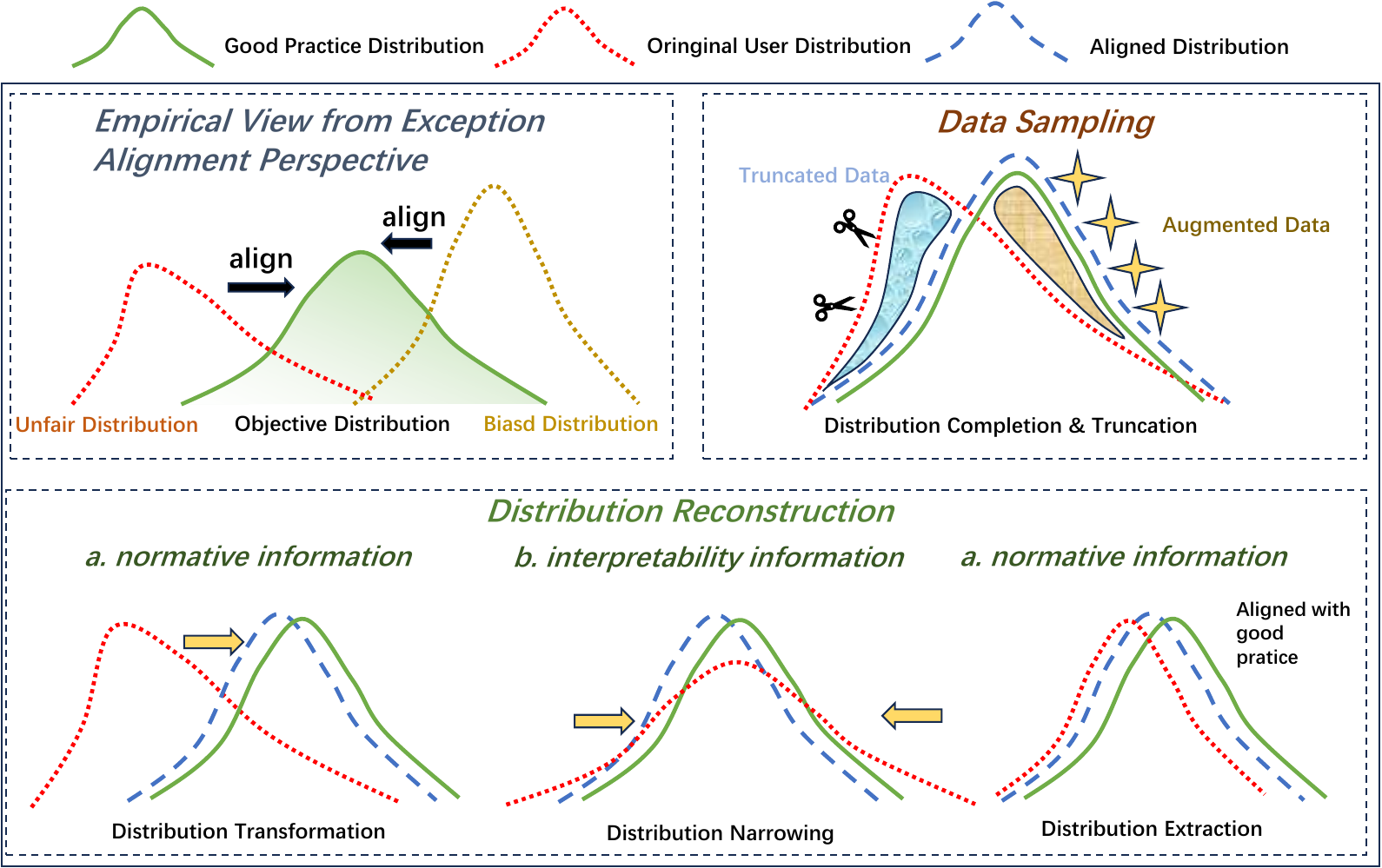}
  \caption{Aligning developers' exception handling from biased, user-oriented practices to industry-standard “good practice” distributions through iterative data refinement. Distribution truncation, augmentation, and reconstruction guide a progression from coarse-grained reminders to fine-grained, scenario-specific guidance—closing the gap between current human methods and stable, high-quality exception handling.}
  \label{fig3}
  \vskip -.1in
\end{figure*}

In this section, we introduce four progressively refined prompting strategies to guide developers—both humans and LLMs—toward stable and generalizable exception handling practices. As shown in Figure\ref{fig3}, our goal is to align developer performance with recognized best practices, gradually helping them move from vague awareness to well-structured and generalizable handling strategies. We term the vulnerable code segments as \textit{Fraile code}, emphasizing that these code fragments are particularly susceptible to runtime exceptions and error propagation if not addressed properly.

Specifically, we present: \textit{Coarse-grained Reminding prompting}, \textit{Fine-grained Reminding prompting}, \textit{Fine-grained Inspiring prompting}, and \textit{Fine-grained Guiding prompting}. Each prompt setting provides incrementally richer contextual information and guidance about exceptions, their types, and their handling strategies. This incremental approach is designed to gradually improve the developer’s in-context learning process, ensuring a more accurate understanding of exceptions and a stable, repeatable handling methodology that can be applied across various development scenarios.

\textbf{Coarse-grained Reminding prompting} simply alerts developers to “pay attention to potential exceptions,” nudging them to identify and handle Fraile code based on their own experience. As shown in Figures~\ref{fig1.1}, \ref{fig2.1}, and \ref{fig2.2}, while such a reminder can make both human and LLM developers more aware of exceptions, it does not significantly improve the precision of identifying Fraile code. Related studies \cite{kpc} have summarized this phenomenon as a series of bad practices—such as \textit{Incorrect exception handling}—where the developer’s initial intuition is insufficient for robust code improvement.

\textbf{Fine-grained Reminding prompting} focuses on specific exception types in the Fraile code scenario, prompting developers to consider their sources and standardize their handling. Although this level of detail encourages consultation of external documents or examples, these references are often too abstract, non-standardized, or insufficiently generalizable. Consequently, developers may still catch exceptions inaccurately, failing to fundamentally mitigate underlying risks. Prior work has identified patterns such as the \textit{Abuse of try-catch} as common pitfalls even under such guidance.

\textbf{Fine-grained Inspiring prompting} goes a step further by providing a code-level scenario analysis for the Fraile code. Here, intuitive and interpretable natural language guidance helps developers gain deeper insight and analytical capability. While recent studies have shown that such prompting can lead to stable good practices for standalone function-level Fraile code, the complexity and dependency chains of real-world scenarios still pose a challenge. As noted by \cite{bugs}, even experienced developers can introduce errors in exception handling within complex projects.

\textbf{Fine-grained Guiding prompting} finally offers a generalized handling strategy for the identified exceptions. Building upon the previously established stable exception detection performance, developers are now equipped with a structured, generalizable approach. Prior recommendations \cite{difficulty1} strongly advocate for employing such strategies, as developers struggle to perform high-quality optimization without fully understanding the nature of the exception type. By establishing a systematic framework for addressing Fraile code, developers—whether human or model-based—are more likely to achieve high-quality exception handling in diverse and complex contexts.

In essence, these four prompt settings represent a progression from broad, coarse-grained reminders to fine-grained, scenario-specific guidance and ultimately to generalizable handling strategies. By gradually improving the developer’s understanding of exceptions and providing actionable insights, we enhance the robustness and quality of the code they produce. This approach highlights the significant influence of prompt specification on LLM-driven code generation and encourages further research into how different levels of guidance can surpass traditional human practice, better aligning the final implementation with recognized exception handling best practices.

Notably, many programming languages offer three main ways to handle exceptions: declaring them with the \texttt{throws} keyword in the method signature, throwing them with the \texttt{throw} keyword inside the method body, and capturing them via \texttt{try-catch} blocks. Existing work \cite{exception1} points out that simply using \texttt{throws} at the method signature may not reflect the true runtime conditions, as these exceptions propagate up the call stack until caught. Similarly, exceptions thrown within a method body must eventually be handled by the caller via \texttt{try-catch}. Thus, \texttt{try-catch} blocks represent the most practical and common approach. In our method, we adopt this third technique as the best practice, integrating it into our prompting strategies to guide developers toward stable and high-quality exception handling.

\subsubsection{Detailed CEE Construction and Refinement Process}
\label{apx:cee_details}

This appendix provides the complete rationale, methodologies, and iterative steps undertaken to build the CEE in detail. It includes:

\paragraph{Comprehensive Documentation from the JDK:}  
We begin with 433 exception nodes drawn from the official Java documentation, spanning 62 branches across five hierarchical layers. Each node is anchored to a standard JDK-defined exception class or interface.

\paragraph{Integration of Enterprise Practices:}  
To enhance the practicality of the CEE, we incorporate patterns and insights drawn from enterprise-level Java development documentation and established open-source projects. By analyzing widely adopted handling practices, logging standards, and fallback mechanisms, we align the CEE with real-world coding scenarios, ensuring that recommended handling logic is both credible and effective.

\paragraph{Granular Structuring via Scenario, Property, and Handling Logic:}  
For each exception node, we record:
\begin{itemize}
    \item \textbf{Scenario}: Typical contexts or operations where the exception may arise.
    \item \textbf{Property}: Attributes such as exception severity, root causes, and environmental factors.
    \item \textbf{Handling Logic}: Recommended strategies, including try-catch patterns, logging techniques, and fallback operations.
\end{itemize}
This granular detailing enables developers and LLMs to map from a given exception scenario to an appropriate handling strategy more accurately.

\paragraph{Reinforcement Learning-Based Fine-Tuning:}  
We employ a testing framework that uses RL-based fine-tuning to improve the mapping between exceptions and handling logic. Over multiple iterations, false positives and negatives in suggested handling methods are identified and rectified, ensuring that the CEE remains both precise and adaptive.

\paragraph{Iterative Refinement and Community Input:}  
The CEE is treated as a living document, continuously refined through user feedback and community contributions. Over time, newly identified exception patterns, handling techniques, or corrections are integrated, ensuring that the CEE evolves alongside prevailing development practices and tooling.

By following these guidelines and iterative enhancements, the CEE aims to serve as a robust, trusted reference that enhances both human and LLM-driven exception management in Java.

\newpage
\onecolumn
\subsubsection{Seeker Framework} \label{ag1}

\begin{algorithm*}
\small
\caption{Seeker Framework}
\KwIn{Codebase $C$}
\KwOut{Optimized code $C'$ with robust exception handling}
Segment the codebase $C$ into manageable units $U = \{u_1, u_2, \dots, u_N\}$\;
\ForEach{code segment $u_i$ in $C$}{
    \If{(length of $u_i$ is within predefined limit) \textbf{and} (function nesting level is low) \textbf{and} (logical flow is clear)}{
        Add $u_i$ to $U$\;
    }
}
Initialize optimized units $U' = \{\}$\;
\ForEach{unit $u_i$ in $U$}{
    \tcp{Detection Phase}
    Initialize potential exception set $E_i = \{\}$\;
    Use the \textbf{Detector} agent to analyze unit $u_i$\;
    \textbf{In parallel do} \{
        \tcp{Static Analysis}
        Generate control flow graph $CFG_i$ and exception propagation graph $EPG_i$ for $u_i$\;
        Identify sensitive code segments $S_{i}^{\text{static}} = \{s_{i1}^{\text{static}}, s_{i2}^{\text{static}}, \dots\}$ in $u_i$\;
        \tcp{Scenario and Property Matching}
        Perform scenario and property matching on $u_i$\;
        Identify sensitive code segments $S_{i}^{\text{match}} = \{s_{i1}^{\text{match}}, s_{i2}^{\text{match}}, \dots\}$ in $u_i$\;
    \}
    Combine sensitive code segments: $S_i = S_{i}^{\text{static}} \cup S_{i}^{\text{match}}$\;
    \ForEach{segment $s_{ij}$ in $S_i$}{
        Detect potential exception branches $E_{bij}$ in $s_{ij}$\;
        $E_{bi} \leftarrow E_{bi} \cup E_{bij}$\;
    }

    \tcp{Retrieval Phase}
    Use the \textbf{Predator} agent to retrieve fragile code and try-catch blocks\;
    Summarize unit $u_i$ at the function level to obtain code summary $F_i$\;
    Perform Deep-RAG using $F_i$ and exception branches $E_{bi}$, get exception nodes $E_{ni}$\;
    Mapping relevant exception handling strategies $H_i = \{h_{i1}, h_{i2}, \dots\}$ from CEE\;
    \tcp{Ranking Phase}
    Use the \textbf{Ranker} agent to assign grades to exceptions in $E_{ni}$\;
    \ForEach{exception $e_{ik}$ in $E_{ni}$}{
        Calculate exception likelihood score $l_{ik}$ based on $e_{ik}$ attribute and impact\;
        Calculate suitability score $u_{ik}$ of handling strategy $h_{ik}$\;
        Compute overall grade $g_{ik} = \alpha \cdot l_{ik} + \beta \cdot u_{ik}$\;
    }
    Rank exceptions in $E_{ni}$ based on grades $g_{ik}$ in descending order to get ranked list $E_{ni}'$\;
    \tcp{Handling Phase}
    Use the \textbf{Handler} agent to generate optimized code $u_i'$\;
    \ForEach{exception $e_{ik}$ of $E_{ni}'$ if $g_{ik} > \gamma$}{
        Mapping handling strategy $h_{ik}$ from $H_i$\;
        Apply $h_{ik}$ to code segment(s) related to $e_{ik}$ in $u_i$\;
    }
    $U' \leftarrow U' \cup \{u_i'\}$\;
}
Combine optimized units $U'$ to produce the final optimized code $C'$\;
\end{algorithm*}

\newpage
\subsubsection{Deep-RAG Algorithm} \label{apx1.1}

\begin{algorithm*}
\caption{Deep Retrieval-Augmented Generation (Deep-RAG)}
\KwIn{Knowledge hierarchy tree $T$, unit summary $F_i$, detected queries $Q_i$, environment context $Env$}
\KwOut{Relevant information retrievals $R_i$}
Initialize relevant knowledge branches set $B = \{\}$\;
Assign knowledge scenario labels $L = \{l_1, l_2, \dots\}$ to branches of $T$\;
\ForEach{query $q_{ik}$ in $Q_i$}{
    Identify branches $B_{ik}$ in $T$ related to $q_{ik}$ based on labels $L$\;
    $B \leftarrow B \cup B_{ik}$\;
}
\ForEach{branch $b_m$ in $B$}{
    \tcp{Verification Step}
    Select few-sample document examples $X_m = \{x_{m1}, x_{m2}, \dots\}$ associated with branch $b_m$\;
    \ForEach{example $x_{mj}$ in $X_m$}{
        Perform query matching to obtain pass rate $p_{mj}$ and capture accuracy $a_{mj}$\;
        \If{$p_{mj}$ or $a_{mj}$ below threshold $\theta$}{
            Record failure pattern $fp_{mj}$ based on $Env$\;
            Update environment context $Env$ with $fp_{mj}$\;
        }
    }
    Compute average pass rate $\bar{p}_m$ and accuracy $\bar{a}_m$ for branch $b_m$\;
    \If{$\bar{p}_m$ or $\bar{a}_m$ below threshold $\theta$}{
        Fine-tune labels $L$ for branch $b_m$ based on aggregated feedback from $Env$\;
    }
}
Initialize information retrievals set $R_i = \{\}$\;
\ForEach{branch $b_m$ in $B$}{
    Select depth level $D$ for node evaluation\;
    \For{$d = 1$ to $D$}{
        \ForEach{node $n_{ml}$ at depth $d$ in branch $b_m$}{
            Evaluate relevance score $r_{ml}$ to summary $F_i$ and queries $Q_i$\;
            \If{$r_{ml} > \delta$}{
                Retrieve information $r_{ml}$ from knowledge base\;
                $R_i \leftarrow R_i \cup \{r_{ml}\}$\;
            }
        }
    }
}
\end{algorithm*}
\twocolumn
In the Deep-RAG algorithm, we assign development scenario labels to each branch of the exception inheritance tree based on their inheritance relationships, enabling the identification of branches that may correspond to specific information of fragile code segments. Acting as an intelligent agent, the algorithm interacts dynamically with its operational environment by leveraging feedback from detection pass rates and capture accuracies obtained during the few-shot verification step. This feedback mechanism allows the system to refine the granularity and descriptions of the scenario labels through regularization prompts derived from failed samples. As a result, Deep-RAG can accurately identify the risk scenarios where fragile codes are located and the corresponding knowledge branches that are activated. Subsequently, the algorithm selectively performs node evaluations on these branches by depth, thereby enhancing retrieval performance and optimizing computational overhead. Additionally, we have designed the algorithm interface to be highly general, ensuring its applicability across a wide range of RAG scenarios beyond exception handling. This generality allows Deep-RAG to support diverse applications, as further detailed in Appendix \ref{apx2.5}. By integrating environmental feedback and maintaining a flexible, agent-based interaction model, Deep-RAG not only improves retrieval accuracy and efficiency but also adapts seamlessly to various domains and information retrieval tasks, demonstrating its versatility and robustness in enhancing the performance of large language models.

\subsection{Experimental Details} \label{apx2}

\subsubsection{Metrics} \label{sec:formulations}
\begin{enumerate}
    \item \textbf{Automated Code Review Score (ACRS)}: This metric evaluates the overall quality of the generated code in terms of adherence to coding standards and best practices, based on an automated code review model. \label{eq:acrs}

    \begin{equation}
    \text{ACRS} = \frac{\sum_{i=1}^{N} w_i s_i}{\sum_{i=1}^{N} w_i} \times 100\%
    \end{equation}
    
    where:
    
    \begin{itemize}
      \item \( N \) is the total number of code quality checks performed by the automated code review tool.
      \item \( w_i \) is the weight assigned to the \( i \)-th code quality rule, reflecting its importance.
      \item \( s_i \) is the score for the \( i \)-th rule, defined as:
        \begin{equation}
        s_i = \frac{q_i}{Q_i}
        \end{equation}
        where:
        \begin{itemize}
          \item \( q_i \) is the raw score for the \( i \)-th rule, based on the specific quality measure (e.g., code readability, efficiency, etc.).
          \item \( Q_i \) is the maximum possible score for the \( i \)-th rule, which ensures that \( s_i \) is normalized to the range \( [0, 1] \).
        \end{itemize}
    \end{itemize}

    A higher ACRS indicates better adherence to coding standards and best practices.

    \item \textbf{Coverage (COV)}: This metric measures the proportion of actual sensitive code segments that our method successfully detects.
    \label{eq:cov}

    Let \( S = \{ s_1, s_2, \dots, s_N \} \) be the set of actual sensitive code segments.

    Let \( D = \{ d_1, d_2, \dots, d_M \} \) be the set of detected sensitive code segments.

    Define an indicator function:

    \[
    I_{\text{detected}}(s_i) = \begin{cases}
    1, & \text{if } \exists d_j \in D \text{ such that } d_j = s_i \\
    0, & \text{otherwise}
    \end{cases}
    \]

    Then, the Coverage is defined as:

    \[
    \text{COV} = \frac{\sum_{i=1}^N I_{\text{detected}}(s_i)}{N} \times 100\%
    \]

    This metric reflects the percentage of actual sensitive code segments correctly detected by our method. Over-detection (detecting more code segments than actual sensitive code) is not penalized in this metric.

    \item \textbf{Coverage Pass (COV-P)}: This metric assesses the accuracy of the try-blocks detected by the \textbf{Predator} agent compared to the actual code that requires try-catch blocks, penalizing over-detection.
    \label{eq:cov-p}
    Let \( T = \{ t_1, t_2, \dots, t_P \} \) be the set of actual code regions that should be enclosed in try-catch blocks (actual try-blocks).

    Let \( \hat{T} = \{ \hat{t}_1, \hat{t}_2, \dots, \hat{t}_Q \} \) be the set of code regions detected by the \textbf{Predator} agent as requiring try-catch blocks (detected try-blocks).

    Define an indicator function:

    \[
    I_{\text{correct}}(\hat{t}_j) = \begin{cases}
    1, & \text{if } \hat{t}_j \in T \\  
    0, & \text{otherwise}
    \end{cases}
    \]

    The number of correctly detected try-blocks is:

    \[
    \text{TP} = \sum_{j=1}^Q I_{\text{correct}}(\hat{t}_j)
    \]

    The number of false positives (incorrectly detected try-blocks) is:

    \[
    \text{FP} = Q - \text{TP}
    \]

    The number of false negatives (actual try-blocks not detected) is:

    \[
    \text{FN} = P - \text{TP}
    \]

    We define the Coverage Pass (COV-P) as:

    \[
    \text{COV-P} = \frac{\text{TP}}{P + \text{FP}} \times 100\%
    \]

    This formulation penalizes over-detection by including the false positives in the denominator. A try-block is considered correct if it exactly matches the actual code lines; any over-marking or under-marking is counted as incorrect.

    \item \textbf{Accuracy (ACC)}: This metric evaluates the correctness of the exception types identified by the \textbf{Predator} agent compared to the actual exception types.
    \label{eq:acc}

    Let \( E = \{ e_1, e_2, \dots, e_R \} \) be the set of actual exception types that should be handled.

    Let \( \hat{E} = \{ \hat{e}_1, \hat{e}_2, \dots, \hat{e}_S \} \) be the set of exception types identified by the \textbf{Predator} agent.

    Define an indicator function:

    \[
    I_{\text{correct}}(\hat{e}_j) = \begin{cases}
    1, & \text{if } \hat{e}_j = e_i \\
    1, & \text{if } \hat{e}_j\text{ is a subclass of } e_i\\
    0, & \text{otherwise}
    \end{cases}
    \]

    Then, the Accuracy is defined as:

    \[
    \text{ACC} = \frac{\sum_{j=1}^S I_{\text{correct}}(\hat{e}_j)}{S} \times 100\%
    \]

    This metric reflects the proportion of identified exception types that are correct, considering subclass relationships. Over-detection of incorrect exception types decreases the accuracy.

    \item \textbf{Edit Similarity (ES)}: This metric computes the text similarity between the generated try-catch blocks and the actual try-catch blocks.
    \label{eq:es}

    Let \( G \) be the generated try-catch code, and \( A \) be the actual try-catch code.

    The Edit Similarity is defined as:

    \[
    \text{ES} = 1 - \frac{\text{LevenshteinDistance}(G, A)}{\max(|G|, |A|)}
    \]

    where \( \text{LevenshteinDistance}(G, A) \) is the minimum number of single-character edits (insertions, deletions, or substitutions) required to change \( G \) into \( A \), and \( |G| \), \( |A| \) are the lengths of \( G \) and \( A \), respectively.

    A higher ES indicates that the generated code closely matches the actual code.

    \item \textbf{Code Review Score (CRS)}: This metric involves submitting the generated try-catch blocks to an LLM-based code reviewer (e.g., GPT-4o) for evaluation. The language model provides a binary assessment: \emph{good} or \emph{bad}.
    \label{eq:crs}

    Let \( N_{\text{good}} \) be the number of generated try-catch blocks evaluated as \emph{good}, and \( N_{\text{total}} \) be the total number of try-catch blocks evaluated.

    The Code Review Score is defined as:

    \[
    \text{CRS} = \frac{N_{\text{good}}}{N_{\text{total}}} \times 100\%
    \]

    This metric reflects the proportion of generated exception handling implementations that are considered good according to engineering best practices.
\end{enumerate}
\begin{table*}[t]
\centering
\caption{The Excerpt Data source}
\label{table:source}
\begin{tabular}{lcccccc}
\toprule
\textbf{Repo} & \textbf{Commits} & \textbf{Stars} & \textbf{Forks} & \textbf{Issue Fix} & \textbf{Doc} & \textbf{Under Maintenance}\\
\midrule
Anki-Android       & 18410  & 8500  & 2200  & 262  & Y & Y \\
AntennaPod      & 6197  & 6300  & 1400  & 295  & Y & Y  \\
connectbot           & 1845  & 2480 & 629  & 321  & N/A & Y \\
FairEmail             & 30259  & 3073  & 640  & N/A  & Y & Y  \\
FBReaderJ        & 7159  & 1832  & 802  & 248  & Y & N/A \\
FP2-Launcher        & 1179  & 25  & 2  & 16 & Y & N/A \\
NewsBlur          & 19603  & 6800  & 995  & 158 & Y & Y \\
Launcher3        & 2932 & 91 & 642 & 2 & N/A & Y \\
Lawnchair-V1      & 4400  & 93  & 43  & 394  & Y & Y \\
MozStumbler      & 1727  & 619  & 212  &  203 & Y & N/A \\
\bottomrule
\end{tabular}
\end{table*}
\subsubsection{Datasets} \label{apx2.1}
To ensure the quality and representativeness of the dataset, we carefully selected projects on GitHub that are both active and large in scale. We applied stringent selection criteria, including the number of stars, forks, and exception handling repair suggestions in the project \citep{difficulty2}, to ensure that the dataset comprehensively covers the exception handling practices of modern open-source projects. By automating the collection of project metadata and commit history through the GitHub API, and manually filtering commit records related to exception handling, we have constructed a high-quality, representative dataset for exception handling that provides a solid foundation for evaluating Seeker.


We quantify the quality of datasets in the context of code generation and exception handling using multiple dimensions, encompassing project popularity, community engagement, codebase quality, security posture, documentation integrity and dynamic maintenance. To provide a holistic assessment, we propose a Composite Quality Metric (CQM) that aggregates these dimensions into a single quantitative indicator. Open source code repositories that perform well under this metric enter our semi-automated review process to screen high-quality exception handling blocks for few-shot, CEE building, or testing.

To avoid data leakage, we also performed a round of variations on the test set. Considering that our method does not directly rely on data but fully utilizes the LLM's ability to understand and reason about code, the evaluation results are consistent with our predictions, and the impact of data leakage on the credibility of our method is negligible.

\onecolumn
\subsubsection{Prompt and Document} \label{apx2.2}

\begin{tcolorbox}[breakable, title=CEE Prompt Template]
genscenario = ``````Below is a kind of exception in java. Please according to the sample
discription of scenario of errortype, provide a scenario description of the
exception in java just like the sample description.Please note that the
granularity of the scenario descriptions you generate should be consistent
with the examples.\\

[Sample Description] \\
\{\textit{sample\_desc}\}
\\

[Exception] \\
\{\textit{ename}\}
\\

Note you should output in the json format like below, please note that the 
granularity of the scenario descriptions you generate should be consistent
with the examples:\\
\{\{\\
$\mbox{\ \ \ \ }$``scenario": ...\\
\}\}\\
"""
\\
genproperty = ``````Below is a kind of exception in java and its scenario description. Please
according to the sample discription of scenario and property of errortype,
provide a property description of the exception in java just like the sample
description. You can alse adjust the given scenario description to make them
consistent. Please note that the granularity of the property descriptions you
generate should be consistent with the examples.\\

[Sample Description]\\
\{\textit{sample\_desc}\}
\\

[Exception]\\
\{\textit{ename}\}
\\

[Scenario Description]\\
\{\textit{scenario}\}
\\

Note you should output in the json format like below, please note that the
granularity of the property descriptions you generate should be consistent
with the examples:\\
\{\{\\
$\mbox{\ \ \ \ }$``scenario": ...;\\
$\mbox{\ \ \ \ }$``property": ...\\
\}\}\\
"""
\end{tcolorbox}

\begin{tcolorbox}[breakable, title = Planner Prompt Template]
planner\_prompt = ``````You are a software engineer tasked with analyzing a codebase. Your task is 
to segment the given codebase into manageable units for further analysis. The 
criteria for segmentation are:
\\
- Each unit should have a length within 200 lines.\\
- The function nesting level should be low.\\
- The logical flow should be clear and self-contained.\\
- The segment should be complete and readable.\\

Given the following codebase:\\

[Codebase]\\
\{\textit{codebase}\}\\

Please segment the codebase into units and list them as:\\

Unit 1:[Code Segment]\\
\{\{unit1\}\}\\

Unit 2:[Code Segment]\\
\{\{unit2\}\}\\
...\\

Ensure that each unit complies with the criteria specified above.\\
"""
\end{tcolorbox}

\begin{tcolorbox}[breakable, title = Detector Prompt Template]
detector\_senario\_match =  ``````You are a java code auditor. You will be given a doc describe 
different exception scenarios and a java code snippet.\\

Your task is to label each line of the code snippet with the exception
scenario that it belongs to. If a line does not belong to any scenario,
label it with ``None". If a line belongs to one of the given scenarios,
label it with all the scenarios it belongs to.\\

[Scenario description]\\
\{\textit{scenario}\}\\

[Java code]\\
\{\textit{code}\}\\

Please output the labeling result in the json format like below:\\
\{\{\\
$\mbox{\ \ \ \ }$``code\_with\_label": ...\\
\}\}\\
"""

detector\_prop\_match = 
``````You are a java code auditor. You will be given a doc describe 
different exception properties and a java code snippet.\\

Your task is to label each line of the code snippet with the exception
property that it belongs to. If a line does not belong to any property,
label it with ``None". If a line belongs to one of the given properties,
label it with all the properties it belongs to.\\

[property description]\\
\{\textit{property}\}\\

[Java code]\\
\{\textit{code}\}\\

Please output the labeling result in the json format like below:\\
\{\{\\
$\mbox{\ \ }$``code\_with\_label": ...\\
\}\}\\
"""
\end{tcolorbox}

\begin{tcolorbox}[breakable, title = Predator Prompt Template]
predator\_prompt = 
``````You are a code analysis assistant. Your task is to process the given
code unit and identify specific exception types that may be thrown.\\

[Code Unit]\\
\{\textit{code\_unit}\}\\

[Code Summary]\\
\{\textit{code\_summary}\}\\

Based on the code summary and the potential exception branches provided,
identify the specific exception nodes that may be thrown.\\

[Potential Exception Branches]\\
\{\textit{exception\_branches}\}\\

Please answer in the following JSON format:\\
\{\{\\
$\mbox{\ \ \ \ }$``ExceptionNodes":$\mbox{\ }$[\\
$\mbox{\ \ \ \ \ \ \ \ }$\{\{\\
$\mbox{\ \ \ \ \ \ \ \ \ \ \ \ }$``ExceptionType":$\mbox{\ }$``ExceptionType1",\\
$\mbox{\ \ \ \ \ \ \ \ }$\}\},\\
$\mbox{\ \ \ \ \ \ \ \ }$\{\{\\
$\mbox{\ \ \ \ \ \ \ \ \ \ \ \ }$``ExceptionType":$\mbox{\ }$``ExceptionType2",\\
$\mbox{\ \ \ \ \ \ \ \ }$\}\},\\
$\mbox{\ \ \ \ \ \ \ \ }$...\\
$\mbox{\ \ \ \ }$]\\
\}\}\\
Ensure that your response strictly follows the specified format.\\
"""
\end{tcolorbox}

\begin{tcolorbox}[breakable, title = Ranker Prompt Template]
ranker\_prompt\ =
``````You\ are\ an\ exception\ ranking\ assistant.\ Your\ task\ is\ to\ assign\ grades
to\ the\ identified\ exceptions\ based\ on\ their\ likelihood\ and\ the\ suitability
of\ their\ handling\ strategies.\\

For\ each\ exception,\ please\ calculate:\\

- Exception\ Likelihood\ Score\ (from\ 0\ to\ 1)\ based\ on\ its\ attributes\ and
impact.\\
- Suitability\ Score\ (from\ 0\ to\ 1)\ of\ the\ proposed\ handling\ strategy.\\

[Identified\ Exceptions\ and\ Handling\ Strategies]\\
\{\textit{exception\_nodes}\}\\

Provide\ your\ calculations\ and\ the\ final\ grades\ in\ the\ following\ JSON\ format:\\
\{\{\\
$\mbox{\ \ \ \ }$``Exceptions":\ [\\
$\mbox{\ \ \ \ \ \ \ \ }$\{\{\\
$\mbox{\ \ \ \ \ \ \ \ \ \ \ \ }$``ExceptionType":\ ``ExceptionType1",\\
$\mbox{\ \ \ \ \ \ \ \ \ \ \ \ }$``LikelihoodScore":\ value,\\
$\mbox{\ \ \ \ \ \ \ \ \ \ \ \ }$``SuitabilityScore":\ value,\\
$\mbox{\ \ \ \ \ \ \ \ }$\}\},\\
$\mbox{\ \ \ \ \ \ \ \ }$...\\
$\mbox{\ \ \ \ }$]\\
\}\}\\

Please\ ensure\ your\ response\ adheres\ to\ the\ specified\ format.\\

"""
\end{tcolorbox}

\begin{tcolorbox}[breakable, title = Handler Prompt Template]
handler\_prompt = 
``````You are a software engineer specializing in exception handling. Your 
task is to optimize the given code unit by applying appropriate exception
handling strategies.\\

[Code Unit]\\
\{\textit{code\_unit}\}\\

[Handling Strategy]\\
\{\textit{strategy1}\}\\

Generate the optimized code with the applied exception handling strategies.\\

Please provide the optimized code in the following format:\\

[Optimized Code]\\
\{\{optimized\_code\}\}\\

Ensure that the code is syntactically correct and adheres to best practices
in exception handling.\\
"""
\end{tcolorbox}

\begin{tcolorbox}[breakable, title = Sample CEE Node]
\{\\
$\mbox{\ \ \ \ }$``name":$\mbox{\ }$``IOException",\\
$\mbox{\ \ \ \ }$``children":$\mbox{\ }$[...],\\
$\mbox{\ \ \ \ }$``info":$\mbox{\ }$\{\\
$\mbox{\ \ \ \ \ \ \ \ }$``definition":$\mbox{\ }$``IOException is a checked exception that is thrown when an input-output operation failed or interrupted. It's a general class of exceptions produced by failed or interrupted I/O operations.",\\
$\mbox{\ \ \ \ \ \ \ \ }$``reasons":$\mbox{\ }$``There are several reasons that could cause an IOException to be thrown. These include: File not found error, when the file required for the operation does not exist; Accessing a locked file, which another thread or process is currently using; The file system is read only and write operation is performed; Network connection closed prematurely; Lack of access rights.",\\
$\mbox{\ \ \ \ \ \ \ \ }$``dangerous\_operations":$\mbox{\ }$``Operations that could typically raise an IOException include: Reading from or writing to a file; Opening a non-existent file; Attempting to open a socket to a non-existent server; Trying to read from a connection after it's been closed; Trying to change the position of a file pointer beyond the size of the file.",\\
$\mbox{\ \ \ \ \ \ \ \ }$``sample\_code":$\mbox{\ }$``String fileName = 'nonexistentfile.txt'; \textbackslash n FileReader fileReader = new FileReader(fileName);",\\
$\mbox{\ \ \ \ \ \ \ \ }$``handle\_code":$\mbox{\ }$``String fileName = 'nonexistentfile.txt'; \textbackslash n try \{ \textbackslash n FileReader fileReader = new FileReader(fileName); \textbackslash n \} catch(IOException ex) \{ \textbackslash n    System.out.println('An error occurred while processing the file ' + fileName); \textbackslash n    ex.printStackTrace(); \textbackslash n \}",\\
$\mbox{\ \ \ \ \ \ \ \ }$``handle\_logic":$\mbox{\ }$``Try the codes attempting to establish connection with a file/stream/network, catch corresponding IOException and report it, output openpath is suggested."\\
$\mbox{\ \ \ \ }$\},\\
$\mbox{\ \ \ \ }$``scenario":$\mbox{\ }$``attempt to read from or write to a file/stream/network connection",\\
$\mbox{\ \ \ \ }$``property":$\mbox{\ }$``There might be an unexpected issue with accessing the file/stream/network due to reasons like the file not being found, the stream being closed, or the network connection being interrupted"\\
\}
\end{tcolorbox}


\twocolumn
\subsubsection{Computation Cost Analysis} \label{apx2.4}
Integrating a comprehensive exception handling mechanism like \textbf{Seeker} introduces potential challenges in computational overhead, especially when dealing with a large number of exception types and complex inheritance relationships. To address this, we designed a high-concurrency interface that keeps the additional computing time overhead constant, regardless of the code volume level. This ensures scalability and controllable complexity when processing any size of codebase.

To evaluate the efficiency of our high-concurrency interface, we conducted experiments on 100 Java code files both before and after implementing parallel processing. For each code file, we executed the exception handling process and recorded the time taken. In the parallelized version, while the processing between different code files remained sequential, the processing within each code file—specifically, the CEE retrieval involving branch and layered processing—was parallelized.

The results are summarized in Table~\ref{table:computation_cost}. After applying parallel processing, the average time per code file was reduced to approximately 19.4 seconds, which is about $\frac{1}{15}$ of the time taken with sequential processing. This significant reduction demonstrates the effectiveness of our parallelization strategy.

\begin{table}[ht]
\centering
\caption{Computation Time Before and After Parallelization}
\label{table:computation_cost}
\resizebox{\columnwidth}{!}{
\begin{tabular}{lcc}
\toprule
\textbf{Processing Method} & \textbf{Average Time per Code File (s)} & \textbf{Speedup Factor} \\
\midrule
Sequential Processing      & 291.0 & 1x \\
Parallel Processing (Seeker) & 19.4  & 15x \\
\bottomrule
\end{tabular}
}
\end{table}

Notably, the size of the code files did not affect the processing time, indicating that our method efficiently handles codebases of varying sizes without compromising on speed. This stability ensures that \textbf{Seeker} can perform consistent and efficient exception handling across any code, making it highly suitable for practical applications.

\subsubsection{Further Results on different LLMs} \label{apx2.6}
We use different open-source (e.g. Code Llama-34B \citep{codellama}, WizardCoder-34B \citep{wizardcoder}, Vicuna-13B \citep{vicuna}) and closed-source(e.g. Claude-2 \citep{claude}, GPT-3-davinci \citep{gpt3}, GPT-3.5-turbo \citep{gpt3.5}, GPT-4-turbo \citep{gpt4}, GPT-4o \citep{gpt4o}) LLMs as the agent's internal model to further analyze models' ability for exception handling. The results are summarized in Table~\ref{table:leaderboard}.

\begin{table}[ht]
\centering
\caption{Performance of Different Models on Exception Handling Code Generation}
\label{table:leaderboard}
\resizebox{\columnwidth}{!}{
\begin{tabular}{lcccccc}
\toprule
\textbf{Model} & \textbf{ACRS} & \textbf{COV (\%)} & \textbf{COV-P (\%)} & \textbf{ACC (\%)} & \textbf{ES} & \textbf{CRS (\%)} \\
\midrule
\multicolumn{7}{c}{\textbf{Open-Source Models}} \\
\midrule
Code Llama-34B      & 0.31  & 37  & 35 & 32  & 0.25 & 34  \\
WizardCoder-34B     & 0.37  & 35  & 31  & 29  & 0.28 & 35  \\
Vicuna-13B           & 0.23  & 15  & 9  & 11  & 0.19 & 26 \\
\midrule
\multicolumn{7}{c}{\textbf{Closed-Source Models}} \\
\midrule
Claude-2           & 0.42  & 64  & 59  & 54  & 0.40 & 54  \\
GPT-3-davinci        & 0.56  & 78  & 68  & 60  & 0.48 & 58  \\
GPT-3.5-turbo       & 0.63  & 79  & 72  & 66  & 0.52 & 71  \\
GPT-4-turbo         & 0.84  & \textbf{91}  & \textbf{83}  & 77  & 0.63 & 89  \\
GPT-4o        & \textbf{0.85} & \textbf{91} & 81 & \textbf{79} & \textbf{0.64} & \textbf{92} \\
\bottomrule
\end{tabular}
}
\end{table}

The performance variations among different models can be explained by:

- \textbf{Pre-training Data}: Models pre-trained on larger and more diverse code datasets (e.g., GPT-4o) have a better understanding of programming constructs and exception handling patterns.

- \textbf{Model Architecture}: Advanced architectures with higher capacities and more layers (e.g., GPT-4) capture complex patterns more effectively.

- \textbf{RAG Performance}: Models that efficiently integrate retrieval-augmented generation, effectively utilizing external knowledge (as in our method), perform better.

- \textbf{Understanding Capability}: Models with superior comprehension abilities can accurately detect sensitive code regions and predict appropriate exception handling strategies.

Open-source models, while valuable, may lack the extensive training data and architectural sophistication of closed-source models, leading to lower performance. Closed-source models like GPT-4o and GPT-4 benefit from advanced training techniques and larger datasets, enabling them to excel in tasks requiring nuanced understanding and generation of code, such as exception handling.

\subsection{Other Applicable Scenarios Analysis} \label{apx2.5}
\begin{figure*}[h]
  \centering
  \includegraphics[width=0.9\linewidth]{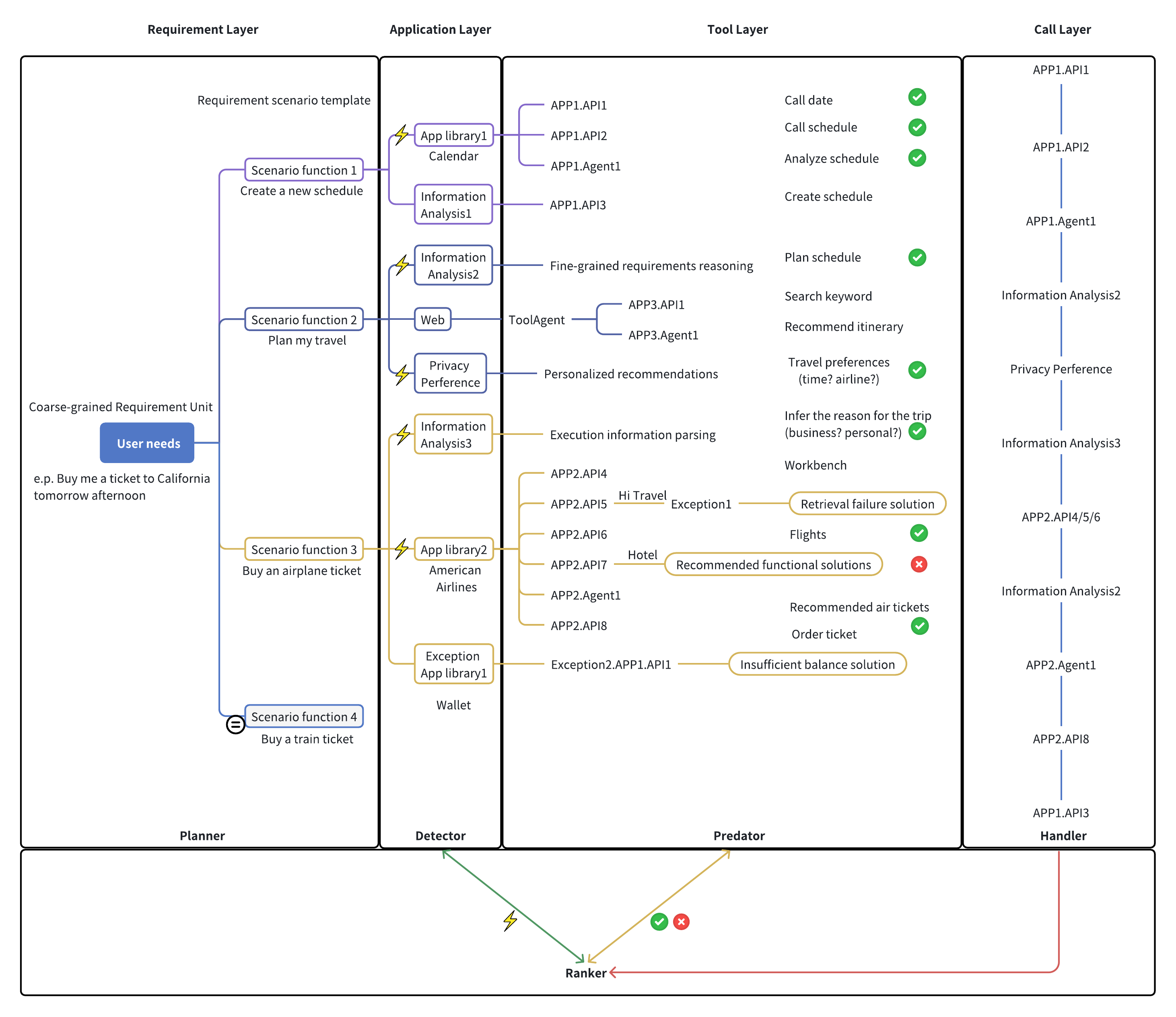}
  \caption{A schematic depiction of integrating the Seeker multi-agent framework into APP requirement engineering workflows. By bridging layered requirements, application functionalities, tool integrations, and call-level operations, Seeker generalizes beyond isolated exception handling to more complex inheritance relationships. This approach improves interpretability, scalability, and reasoning capabilities, demonstrating the framework’s adaptability and high performance across diverse, real-world engineering scenarios.}
  \label{fig8.5}
  \vskip -.1in
\end{figure*}

Figure \ref{fig8.5} shows the migration application of Seeker multi-agent framework in APP requirement engineering that also includes parent-child inheritance relationship. We have reason to believe that Seeker framework can try to be compatible with more complex inheritance relationship, being responsible for reasoning representation, while having high performance and interpretability. The above achievements are not easy to accomplish based on graphs or traditional algorithms.

To validate the general applicability of our system in diverse scenarios, we evaluated \textbf{Seeker} on standard code generation benchmarks, including \textbf{SWE-bench} and \textbf{CoderEval}. We present comparative results demonstrating the incremental improvements achieved by our method.

\textbf{SWE-bench} is an evaluation framework comprising 2,294 software engineering problems derived from real GitHub issues and corresponding pull requests across 12 popular Python repositories\citep{swebench}. It challenges language models to edit a given codebase to resolve specified issues, often requiring understanding and coordinating changes across multiple functions, classes, and files simultaneously. This goes beyond traditional code generation tasks, demanding interaction with execution environments, handling extremely long contexts, and performing complex reasoning.

For our experiments, we selected 50 issues related to exception handling from the SWE-bench Lite dataset. Using \textbf{GPT-4o} as the internal large model, the \textbf{SweAgent}\citep{yang2024sweagentagentcomputerinterfacesenable} coupled with GPT-4o achieved a \textbf{19\%} \emph{resolve rate} and a \textbf{43\%} \emph{apply rate}. In contrast, our \textbf{Seeker} framework attained a \textbf{26\%} resolve rate and a \textbf{61\%} apply rate, indicating a significant improvement.

\begin{table}
\centering
\caption{Performance on SWE-bench Lite Exception Handling Issues}
\label{table:swebench_results}
\resizebox{\columnwidth}{!}{
\begin{tabular}{lcc}
\toprule
\textbf{Method} & \textbf{Resolve Rate (\%)} & \textbf{Apply Rate (\%)} \\
\midrule
SweAgent + GPT-4o & 19 & 43 \\
\textbf{Seeker} + GPT-4o & \textbf{26} & \textbf{61} \\
\bottomrule
\end{tabular}
}
\end{table}

\textbf{CoderEval} is a benchmark designed to assess the performance of models on pragmatic code generation tasks, moving beyond generating standalone functions to handling code that invokes or accesses custom functions and libraries\citep{codereval}. It evaluates a model's ability to generate functional code in real-world settings, similar to open-source or proprietary projects.

In the Java code generation tasks on CoderEval, using \textbf{Codex}\citep{codex} directly yielded a \textbf{Pass@1} score of \textbf{27.83\%}. When integrating our \textbf{Seeker} framework with Codex, the Pass@1 score increased to \textbf{38.16\%}, demonstrating a substantial enhancement in code generation performance.

\begin{table}[h]
\centering
\caption{Performance on CoderEval Java Code Generation Tasks}
\label{table:codereval_results}
\begin{tabular}{lc}
\toprule
\textbf{Method} & \textbf{Pass@1 (\%)} \\
\midrule
Codex & 27.83 \\
\textbf{Seeker} + Codex & \textbf{38.16} \\
\bottomrule
\end{tabular}
\end{table}

These experiments conclusively demonstrate that our \textbf{Seeker} framework can achieve significant incremental improvements across different scenarios and benchmarks. By effectively handling exception-related tasks and enhancing code robustness, \textbf{Seeker} proves to be a valuable addition to existing code generation models, improving their practical applicability in real-world software engineering problems.

Inspired by OpenAI o1 \citep{o1} and DoT \citep{dot}, we found that Seeker framework has more room for development in LLM reasoning. Through pre-deduction in tree inference, LLM is expected to enter the problem-solving ideas more efficiently and optimize its reasoning actions through interaction with the external environment. In the future, we will continue to explore research in this direction.

\section{Related Work}

At present, machine learning has been widely integrated in the field of software engineering, especially in code generation tasks. In this section, we will discuss the progress of Seeker-related work from the latest progress of automatic exception handling tools. These methods have contributed to the robustness or productivity of software engineering, but they also have limitations, which is also the focus of Seeker.

\subsection{Automatic Exception Handling Tools} \label{rw}

Recent work\cite{baseline1} introduced a neural network approach for automated exception handling in Java, which predicts try block locations and generates complete catch blocks in relatively high accuracy. However, the approach is limited to Java and not generalize well without retraining. Additionally, the reliance on GitHub data could introduce biases based on the types of projects and code quality present in the dataset.

SCG\cite{baseline2} conducted an exploratory study on fine-tuning LLM for secure code generation. Their results showed that after fine-tuning issue fixing commits, the secure code generation rate was slightly improved. The best performance was achieved by fine-tuning using function-level and block-level datasets. However, the limitation of this study is still generalization, not directly applicable to other languages. In addition, it limits the amount and the domain of code that can be effectively processed. Little much code beyond training data scale will affect the processing effect. Besides, in terms of automatic vulnerability detection, the use of traditional fine-tuning methods may not fully utilize the domain knowledge in the pre-trained language model, and may overfit to a specific dataset, resulting in misclassification, excessive false positives and false negatives\cite{final}. Its performance is not as good as emerging methods such as prompt-based learning.

Knowledge-driven Prompt Chaining (KPC)\cite{kpc}, an approach to improve code generation by chaining fine-grained knowledge-driven prompts. Their evaluation with 3,079 code generation tasks from Java API documentation showed improvements in exception handling. However, the approach's efficiency relies heavily on the inquiry about built-in exceptions for each built-in JDK, and its practical application is limited if the codebase is complex.

FuzzyCatch\cite{baseline3}, a tool for recommending exception handling code for Android Studio based on fuzzy logic. However, the performance of FuzzyCatch depends on the quality and relevance of the training data. In addition, the tool does not perform well for less common exceptions or domains that are not well represented in the training data.

Neurex\citep{codebert}, a learning-based exception handling recommender that leverages the CodeBERT model to suggest appropriate try-catch blocks, the statements to include within try blocks, and the exception types to catch. However, Neurex still has several limitations. It cannot generate new exception types that were not in the training corpus with low cost. It does not support the generation of exception handling code inside the catch body. Each project might have a different way to handle exception types in the catch body. And Neurex also needs training data, thus, does not work for a new library without any API usage yet. Most importantly, we compared the experimental results and found that even in the experimental granularity of their method, they perform averagely and are primarily good at finding existing exception handling bugs, which is not our focus.
Above all, we have had similar method for baseline so we did not compare with them in the formal experimental part.

A common limitation of these studies is that the training data they rely on may not fully represent all possible coding scenarios. This may result in a model that is effective in specific situations, but may not generalize well to other situations. In addition, the complexity of exception handling in real-world applications may exceed the capabilities of models trained on more common or simpler cases, so it is crucial to call on the understanding and reasoning capabilities of the model itself. The interpretability of exception handling also provides a guarantee for the improvement of developers' programming literacy. The comparison between the above methods and Seeker is shown in figure \ref{fig8}.

\subsection{Multi-agent Collaberation}
Multi-agent collaboration refers to the coordination and collaboration between multiple artificial intelligence (AI) systems, or the symbiotic collaboration between AI and humans, working together to achieve a common goal \citep{mind}. This direction has been explored for quite some time \citep{dy} \citep{book}. Recent developments show that multi-agent collaboration techniques are being used to go beyond the limitations of LLM, which is a promising trajectory. There are many ways for multi-agents to collaborate with LLM.

VisualGPT \citep{visual} and HuggingGPT \citep{hug} explored the collaboration between LLM and other models. Specifically, LLM was used as a decision center to control and call other models to handle more domains, such as vision, speech, and signals. CAMEL \citep{camel} explored the possibility of interaction between two LLMs. These studies mainly use case studies in the experimental stage to demonstrate their effectiveness and provide specific hints for each case.

For multi-agent collaborative software engineering, which is most relevant to Seeker, \cite{dong} introduces quantitative analysis to evaluate agent collaborative code generation. It introduces the waterfall model in software development methods into the collaboration between LLMs. However, there is still a gap between the evaluation benchmarks used and the actual software development scenarios. In addition, although this work builds a fully autonomous system, adding a small amount of guidance from human experts to supervise the operation of the virtual team will help improve the practicality of the method in actual application scenarios. These problems are exactly what we have improved on Seeker.

CODEAGENT\cite{zhang} formalized the repo-level code generation task and proposed a new agent framework based on LLM. CODEAGENT developed five programming tools to enable LLM to interact with software artifacts and designed four agent strategies to optimize the use of tools. The experiment achieved improvements on various programming tasks. However, it only integrated simple tools into CODEAGENT. Some advanced programming tools were not explored. This limitation limits the ability of the agent in some challenging scenarios, such as exception handling tasks.

Above all, nowadays, most code-agent works focus on the transformation from the requirements to code and overlook the code robustness during software evolution, which requires not only understanding the requirement but also dealing with potential exceptions.

\subsection{Robust Software Development Mechanism}
Code robustness refers to the practices and mechanisms that ensure software to run as expected without causing unexpected side effects, security vulnerabilities, or errors. It involves techniques such as type safety, memory safety, and ensuring that all code paths are well-defined, including when exceptions exist. Exception handling is a necessary programming mechanism to maintain code robustness, allowing programs to manage and respond to runtime errors or other abnormal events. It helps maintain the normal flow of execution and ensures that resources are properly released even when errors occur. Exception handling is critical to code robustness because it ensures that unexpected errors do not compromise the stability or security of the system, prevents resource leaks, ensures data integrity, and keeps the program running correctly even when unforeseen errors occur\citep{exception2}. 

From the perspective of code robustness, the defect repair work in the field of software engineering is closely related to exception handling mechanisms, because exception handling involves solving potential errors in the program flow, and developers can mitigate or eliminate defects that may cause program failures or unpredictable behavior\citep{exception3}. Currently, since each defect represents a potential vulnerability or instability in the software and is directly related to the functional correctness of the program, research focuses more on defect repair\cite{v1}, Devign \citep{v1}, VulAdisor \citep{v1}, while the program's exception safety and exception handling, the powerful program defense mechanisms are not considered.

When a program lacks good exception handling, errors may propagate uncontrollably, leading to resource leakage, data corruption, and potential security vulnerabilities. This situation is called fragile code. After the error occurs, Automatic Program Repair related work performs post-processing to fix the code bug\cite{a1}. Representative works include Magis \citep{a2}, PatchFinder \citep{a4}. However, they lack the ability to perceive and repair program risks in advance, and there is a risk of accidentally changing the original function of the code\cite{a3}. 

\newpage
\onecolumn
\begin{figure}[t]
  \centering
  \includegraphics[width=0.85\linewidth]{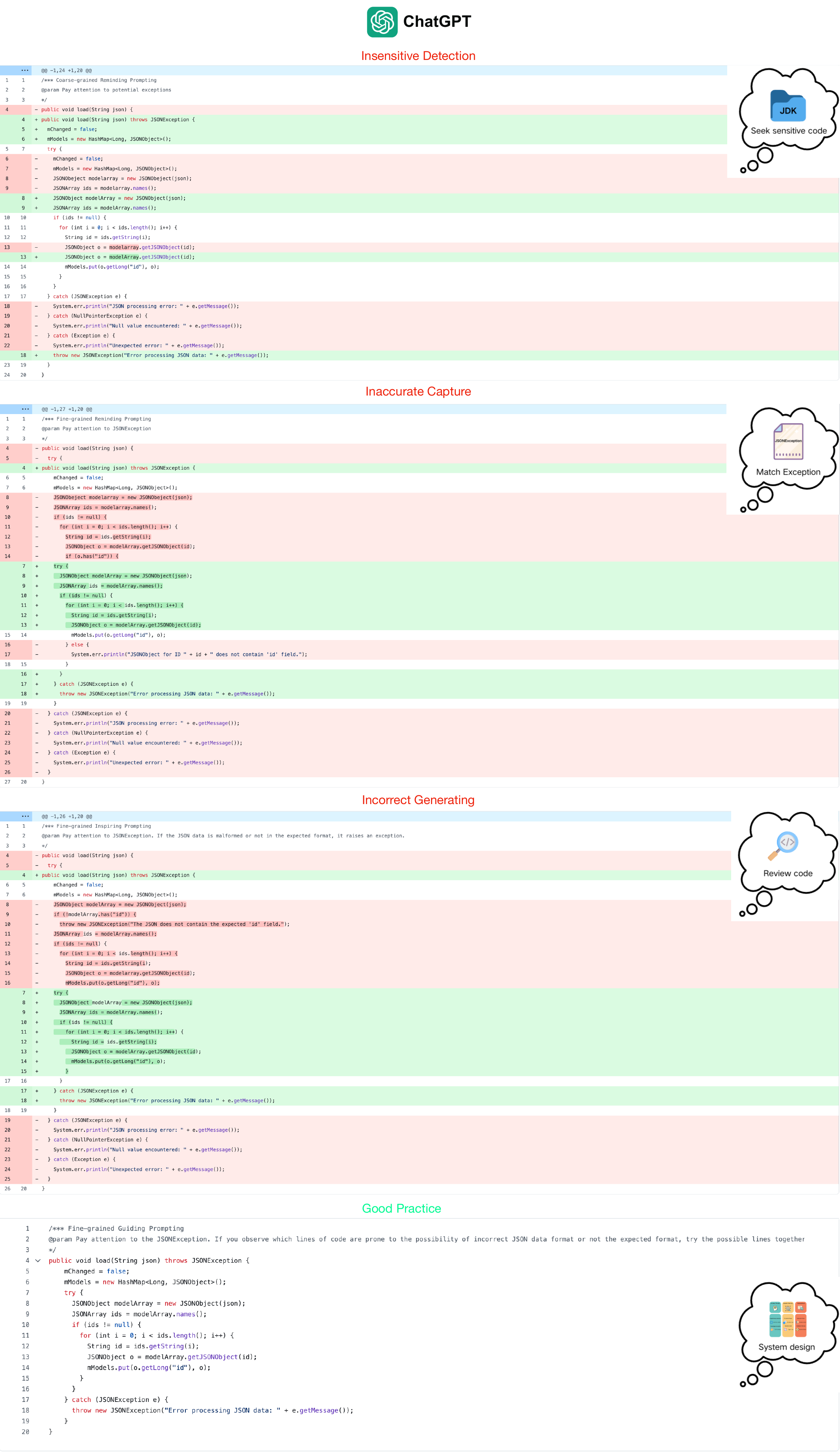}
  \caption{A schematic illustration of the preliminary phenomenon, showing how incremental, targeted guidance enhances LLM-based exception handling. The depicted code segments and annotations highlight which specific information supports more accurate detection and handling of fragile code scenarios.}
  \label{fig2.1}
  \vskip -.1in
\end{figure}
\newpage

\begin{figure}[t]
  \centering
  \includegraphics[width=0.99\linewidth]{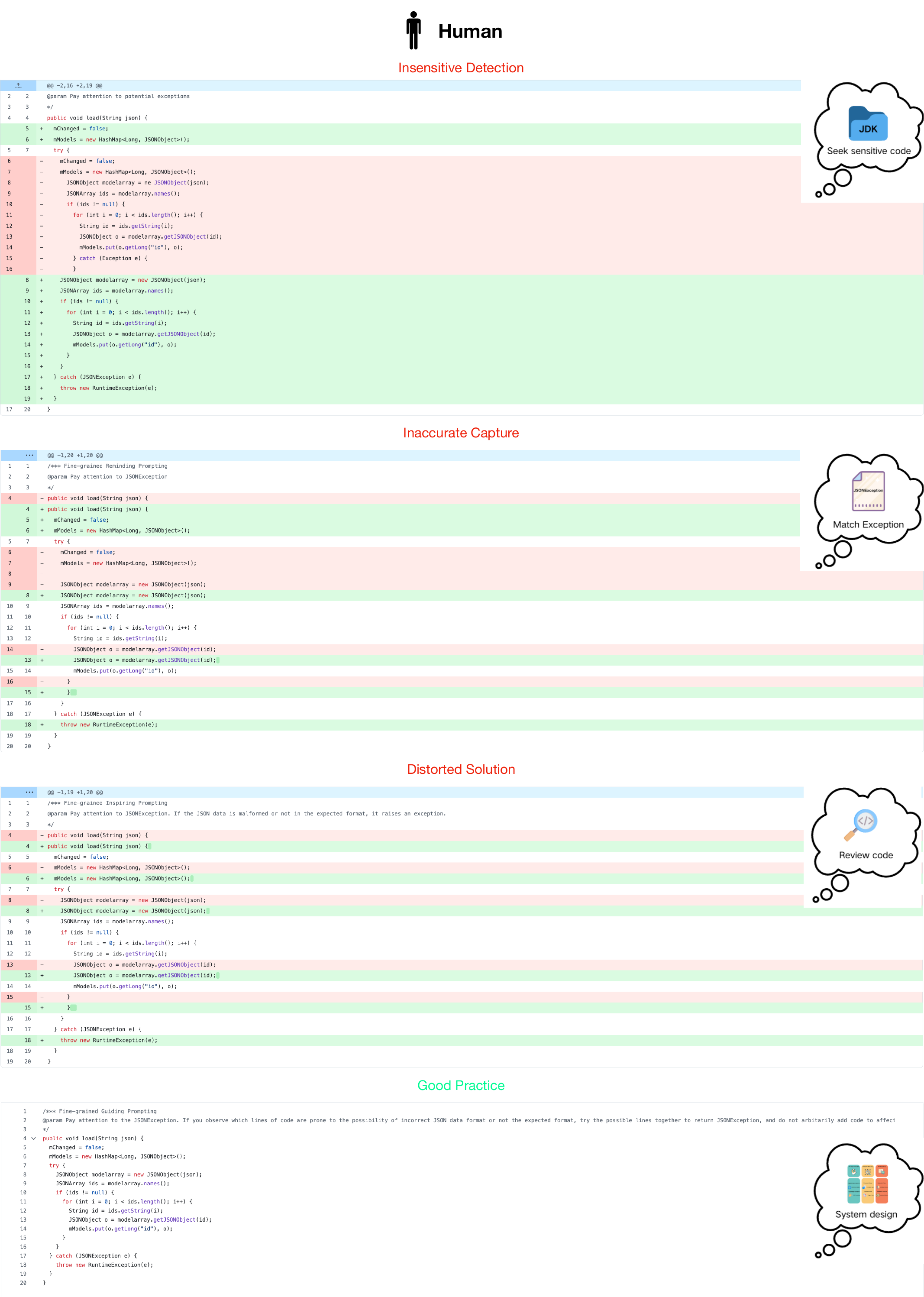}
  \caption{A schematic illustration of the preliminary phenomenon, demonstrating that incremental, targeted guidance similarly benefits both LLMs and human developers in exception handling. The highlighted case study underscores which information elements help bridge the gap between current human practice and reliable, automated handling strategies.}
  \label{fig2.2}
  \vskip -.1in
\end{figure}

\end{document}